\documentclass[letterpaper]{article} 
\usepackage{aaai2026}  
\usepackage{times} 
\usepackage{helvet} 
\usepackage{courier} 
\usepackage[hyphens]{url}  
\usepackage{graphicx} 
\usepackage{natbib}  
\usepackage{caption} 
\usepackage{xcolor} 
\usepackage{subfiles} 
\usepackage{algorithm}
\usepackage{algorithmic}
\usepackage{framed}
\usepackage{booktabs}
\usepackage{color}
\usepackage{amssymb} 
\usepackage{amsmath}
\usepackage{xspace}  
\usepackage{color, colortbl}

\definecolor{lightgray}{rgb}{.91,.91,.91}
\definecolor{rouse}{rgb}{0.981,0.961,0.941}
\definecolor{deepred}{rgb}{0.698,0.133,0.133}
\definecolor{blue}{rgb}{0,0,1}

\usepackage{bbding}
\usepackage{pifont}
\usepackage{amssymb}
\newcommand{\xmarkg}{\textcolor{gray}{\ding{55}}\xspace}
\newcommand{\cmark}{\ding{51}}

\usepackage{newfloat}
\usepackage{listings}
\DeclareCaptionStyle{ruled}{labelfont=normalfont,labelsep=colon,strut=off} 
\floatstyle{ruled}
\newfloat{listing}{tb}{lst}{}
\floatname{listing}{Listing}

\title{SeqWalker: Sequential-Horizon Vision-and-Language Navigation with Hierarchical Planning}
\author{
    Zebin Han\textsuperscript{\rm 1,2,3}, Xudong Wang\textsuperscript{\rm 2,4}\thanks{Corresponding Author}, Baichen Liu\textsuperscript{\rm 2}, Qi Lyu\textsuperscript{\rm 2,4}, Zhenduo Shang\textsuperscript{\rm 2,4}, Jiahua Dong\textsuperscript{\rm 5}, \\  Lianqing Liu\textsuperscript{\rm 2}, Zhi Han\textsuperscript{\rm 2}
}
\affiliations{
    \textsuperscript{\rm 1}North University of China
    \textsuperscript{\rm 2}State Key Laboratory of Robotics and Intelligent Systems, Shenyang Institute of Automation, Chinese Academy of Sciences
    \textsuperscript{\rm 3}Southeast University
    \textsuperscript{\rm 4}University of Chinese Academy of Sciences
    \textsuperscript{\rm 5}Mohamed bin Zayed University of Artificial Intelligence
     \{zebinhan04,dongjiahua1995\}@gmail.com, \{wangxudong,liubaichen,lvqi,shangzhenduo,lqliu,hanzhi\}@sia.cn

}

\usepackage{bibentry}

\begin{document}

\maketitle

\begin{abstract}
{\fontsize{9pt}{11pt}\selectfont
  Sequential-Horizon Vision-and-Language Navigation (SH-VLN) presents a challenging scenario where agents should sequentially execute multi-task navigation guided by complex, long-horizon language instructions. Current vision-and-language navigation models exhibit significant performance degradation with such multi-task instructions, as information overload impairs the agent's ability to attend to observationally relevant details. To address this problem, we propose SeqWalker, a navigation model built on a hierarchical planning framework. Our SeqWalker features: i) A High-Level Planner that dynamically selects global instructions into contextually relevant sub-instructions based on the agent's current visual observations, thus reducing cognitive load; ii) A Low-Level Planner incorporating an Exploration-Verification strategy that leverages the inherent logical structure of instructions for trajectory error correction. To evaluate SH-VLN performance, we also extend the IVLN dataset and establish a new benchmark. Extensive experiments are performed to demonstrate the superiority of the proposed SeqWalker. 
}
\end{abstract}

\begin{links}
\link{Code}{https://seqwalker.github.io/seqwalker/}
\end{links}

\section{Introduction}

Vision-and-Language Navigation (VLN) has emerged as a critical component for embodied AI, requiring intelligent agents to interpret natural language instructions and navigate through complex environments \cite{StreamVLN, MMVLN1, MMVLN2}. This capability holds significant promise for diverse applications, such as autonomous robotics \cite{autorobot1, ICIRA1}, home robots \cite{smart1}, and robots operating in complex dynamic environments \cite{dong2025bring, ICIRA2, guan2025freqspatnet}. While recent advances \cite{NWM, OverVLN} have demonstrated remarkable progress in single-trajectory VLN scenarios, the more practical challenge of sequential multi-task navigation remains largely unaddressed.

In practical applications, agents are often required to execute multiple sub-tasks sequentially within a single deployment session \cite{lmvln, LHVLN2}. For example, on an usual morning, a student asks an agent to help him get up and go to school. This requires the agent to complete a series of tasks in sequence: i) take the clothes from the wardrobe, ii) take me to the bathroom for washing up, iii) prepare breakfast, iv) put the umbrella in my schoolbag, as illustrated in Fig.\ref{fig_1}. The navigation agent is required to complete the user's tasks sequentially and reach the destination. Different from the traditional VLN task, where the user describes one navigation trajectory with one instruction, the user only needs to provide the agent with one sequential-horizon instruction to enable the agent to complete multiple navigation tasks. This is more time-effective for the user, but it also poses more challenges. The sequential-horizon navigation means \textit{larger navigation scenes} with \textit{more complex user instructions}. Current VLN methods \cite{shvln,FindingDory} are dedicated to persistent and efficient navigation in large scenes. Some methods \cite{uninavid,navillm} store the historical state of previous processes and feed it into the agent's current state. Iterative Vision-and-Language Navigation (IVLN) \cite{IVLN} proposes a map-based navigation method that maps the scene to maps during navigation. These advancements improve the navigation performance in persistent scenes and bring VLN research closer to large-scale scene applications. On the other hand, as the navigation scene expands, the complexity of the navigation task increases further, which means that the user describes multiple navigation tasks with one instruction, and the user inevitably describes multi-task trajectories with more complex and longer sequential-horizon instructions. However, the performance of existing VLN models is severely degraded with sequential-horizon instructions, as the excessive information hinders the agent from focusing on details relevant to its current observations.

\begin{figure*}[!t]
\centering
\includegraphics[width=1.0\linewidth]{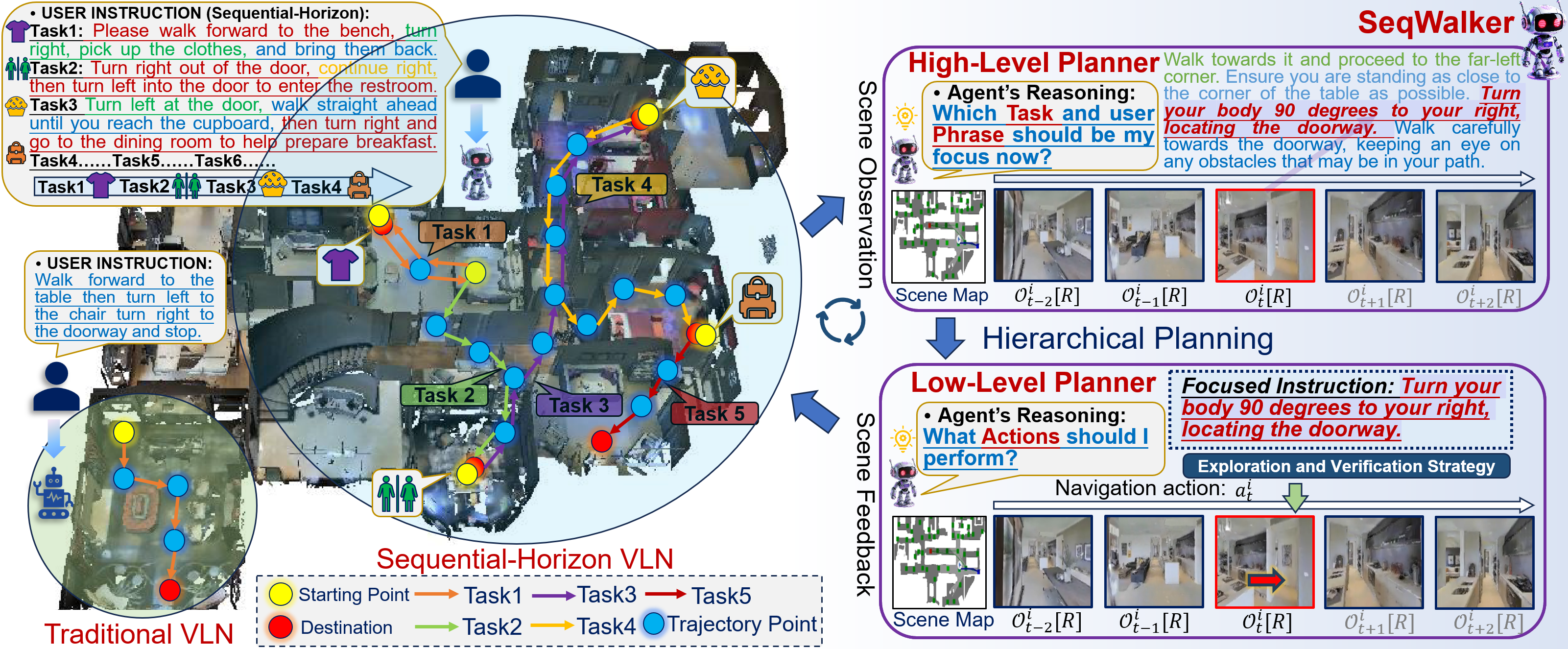}
\caption{Illustration of the proposed \textit{Sequential-Horizon Vision-and-Language Navigation} (SH-VLN) task and our SeqWalker model. Different from traditional VLN, SH-VLN requires navigation agents to follow a sequential multi-task trajectory navigation, while users tend to provide complex long instructions, posing greater challenges. Our SeqWalker adopts a \textit{Hierarchical Planning} strategy, where the \textit{High-Level Planner} selects sub-tasks and instruction phrases based on the agent's observations, and the \textit{Low-Level Planner} provides actions for robust navigation using a proposed Exploration and Verification strategy.
}
\label{fig_1}
\vspace{-4mm}
\end{figure*}

To enable sequential multi-task navigation in large scenes, we introduce a new VLN task called \textit{\underline{S}equential-\underline{H}orizon \underline{V}ision-and-\underline{L}anguage \underline{N}avigation} (SH-VLN). This task emphasizes navigating long distances in large scenes using sequential-horizon instructions, which presents three main challenges: \textit{\textbf{i) Scene Persistence.}} In real-world applications, a deployed navigation agent often operates within the same environment over extended periods \cite{dualmap, r2c}. To ensure efficient operation, the agent should be able to save scene information rather than treat each episode as a new exploration. \textit{\textbf{ii) Complex Long Instructions.}} Different from a single-period navigation task, describing multi-task navigation in large scenes inevitably requires more complex and longer instructions to capture navigation trajectories. However, long and complex instructions interfere with the traditional navigation agents’ understanding of instructions and make it difficult to match them to the current observation, leading to navigation failures. \textit{\textbf{iii) Multi-Trajectory Robustness.}} Sequential-horizon navigation often means longer navigation trajectories. Sequential navigation in long trajectories is more likely to make errors, and once an error occurs, the trajectory is likely to worsen with it, so there are high demands on the robustness of navigation. An intuitive way to solve the long instruction problem is to directly leverage the powerful long sequence reasoning capability of large language models (LLMs). However, directly applying LLMs is ineffective. Although LLMs \cite{gpt4, VLMT} with several hundred billion parameters handle long sequences well, and models with smaller parameters that can be embedded in robots have shortcomings for long sequences. Additionally, directly applying LLM cannot guide accurate actions due to the lack of specific knowledge about the navigation scene.

To solve the SH-VLN task, we propose a SeqWalker, a novel VLN model built on a hierarchical planning framework. SeqWalker is capable of hierarchically understanding the user sequential long instructions step-by-step, while explicitly saving the scene maps to achieve efficient sequential-horizon navigation in large scenes, as shown in the Fig.\ref{fig_1}. Specifically, in \textit{High-Level Planner}, SeqWalker performs local segmentation on global long instructions using pre-trained CLIP models to obtain the most relevant segmented instruction with the agent's current scenes. Compared to global long instructions, the segmentation instructions not only reduce the difficulty of focusing information but also enable corrections for navigation trajectory errors. In \textit{Low-Level Planner}, we develop an Exploration and Verification strategy to achieve forward progress toward destinations or to leverage the inherent logical order of instructions to dynamically correct the current navigation trajectories, which further improves the robustness of long navigation. To evaluate the proposed SH-VLN task, we extend the existing IVLN dataset \cite{IVLN} and propose a new benchmark. Extensive experiments are performed to demonstrate the effectiveness and superiority of our proposed SeqWalker. 

The main contributions of this work are as follows:
\begin{itemize}
    \item We introduce Sequential-Horizon Vision-and-Language Navigation (SH-VLN), a new VLN task where agents follow sequential-horizon instructions to complete a sequential multi-task navigation trajectory. We also provide a new benchmark for evaluating the SH-VLN task.

    \item We propose a SeqWalker model, which is built on a hierarchical planning framework. Our SeqWalker performs local segmentation on long instructions to understand instructions step-by-step while explicitly saving scene maps for sequential-horizon navigation in large scenes. 

    \item We develop an Exploration and Verification strategy. Based on this, SeqWalker navigation in two modes, i.e., forward to destination and trajectory correction, further improving sequential-horizon navigation robustness.

    \item Extensive experiments show that SeqWalker achieves state-of-the-art performance on both traditional IVLN datasets and the proposed sequential-horizon datasets.
\end{itemize}

\begin{figure*}[t!]
\centering
\includegraphics[width=1\linewidth]{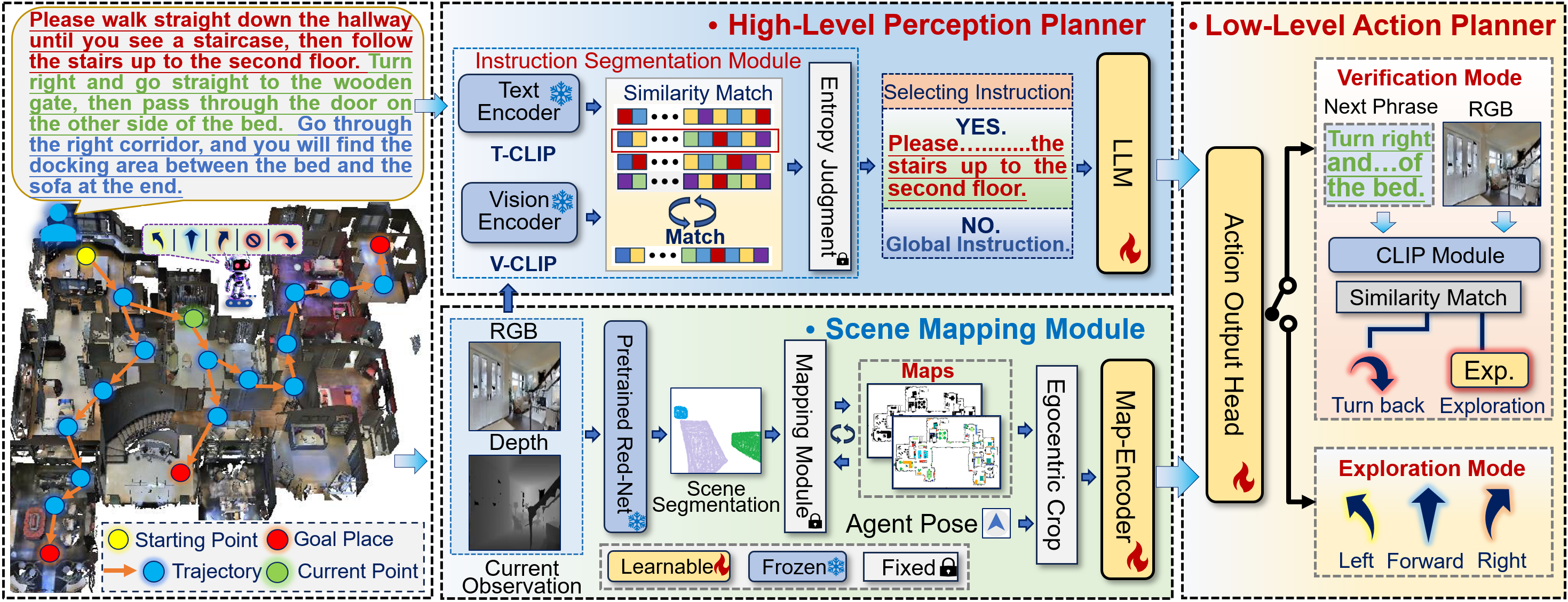}
\caption{Illustration of the proposed SeqWalker pipeline. At each time step $t$, the SeqWalker agent receives RGB and depth observations to follow user instructions. SeqWalker has a hierarchical planning framework. It comprises a \textit{High-Level Perception Planner}, which segments sequential-horizon instructions to obtain the most relevant segmented instruction with the current states. And it also comprises a \textit{Low-Level Action Planner}, which employs an exploration-verification strategy to achieve progress toward destinations or leverage the inherent logical order of instructions to correct navigation trajectories dynamically.
}
\label{fig_2}
\vspace{-3mm}
\end{figure*}

\section{Related Works}
\textbf{Vision-and-Language Navigation:} The task of VLN \cite{liu2025embodied, vln20251,VLNCE3,cite3} is to enable an agent to understand user instructions and navigate in complex environments independently. VLN tasks are typically divided into two types based on the environment: discrete \cite{disvln1,gridmm,uniyolo} and continuous \cite{convln1,convln2,cite4}. In discrete environments, the space is modeled as a graph of connected nodes. The agent moves between these nodes following the given instructions, aiming to reach a target node within a set number of steps. In continuous environments, the agent can move freely using low-level actions such as moving forward or turning, offering more flexibility and a closer resemblance to real-world scenes. Since continuous environments align better with real applications, we follow the continuous VLN setting.

\noindent \textbf{Navigation in Persistent Scenes:} In real-world applications, navigation agents often operate in the same environment for extended periods \cite{r2c,zhang2024navhint, zheng2025esceme,CASM}. Meanwhile, advancements in computing power lead to richer scene datasets \cite{dataset3, MP3D, Habitat3D} and improved simulator platforms \cite{SP1,Habitat}. These developments draw attention to the challenge of navigation in persistent scenes. Some studies \cite{SSVLN,NaVid} address persistent navigation by embedding information from past navigation episodes. However, merely storing scene information as hidden embeddings limits effectiveness. To improve long-term navigation, Iterative VLN (IVLN) task \cite{IVLN} is introduced. IVLN uses a map-based method where it creates semantic maps of the scene during navigation and replaces the original image inputs with these maps. Although these advances bring VLN research closer to large-scene applications, the performance of curren VLN models drops significantly when handling multi-task instructions. We propose SH-VLN to tackle the challenges posed by sequential long-horizon instructions.

\vspace{-1mm}
\section{Problem Formulation}

In VLN-CE task, an agent navigates a continuous scene $\Gamma$ by following user instruction $\mathcal{I}$. In IVLN setting, a large scene $\Gamma$ contains multiple trajectories $\mathcal{T}=\{T_0, T_1,\dots,T_m\}$, and the agent persistently operates within $\Gamma$. In our SH-VLN task, instruction $\mathcal{I}$ is further extended to describe sequential-horizon trajectories $\mathcal{I} \Leftrightarrow \{T_0, T_1,\dots,T_n\}$. Specifically, at each episode $i$, the user provides instruction $\mathcal{I}^{i}$ describing a sequential multi-task trajectory $\{T^i_0, T^i_1,\dots,T^i_n\}$ in $\Gamma$, where $\mathcal{I}^{i}$ consists of phrases $\{S^{i}_0, S^{i}_1, \dots, S^{i}_n\}$. At each time step $t$, the agent predicts the next navigation mode Exploration $\mathcal{E}^{i}_{t}$ or Verification $\mathcal{V}^{i}_{t}$ based on current observation $\mathcal{O}^{i}_{t}$. Mode $\mathcal{E}^{i}_{t}$ aims to move toward the destination, while $\mathcal{V}^{i}_{t}$ corrects trajectory errors. After multiple iterations, the agent reaches destination $\mathcal{G}^{i}$ with a sequential-horizon trajectory. Different from the previous LH-VLN task \cite{lmvln}, SH-VLN builds upon IVLN to evaluate efficient agent operation in persistent scenes. Beyond observation $\mathcal{O}^{i}$ and instruction $\mathcal{I}^{i}$, the agent also stores and leverages scene maps $\mathcal{M}^i$ from previous trajectories $\{T^i_0, T^i_1,\dots,T^i_n\}$ in $\Gamma$ for efficient navigation. Following \cite{VLNCE, IVLN}, agent actions within $\mathcal{E}^i_t$ are defined as $\mathcal{A}$ = \{FORWARD (0.25m), TURN LEFT (15°), TURN RIGHT (15°), STOP\}.

\section{Proposed Methods}

The pipeline of the proposed SeqWalker is illustrated in Fig.\ref{fig_2}, and the overall three navigation steps are as follows:

\textbf{Step I. High-Level Perception Planner}. SeqWalker agent matches the RGB $\mathcal{R}^{i}_{t}$ with the instruction $\mathcal{I}^i$, and selects out the phrase $S^{i}_j$ most suitable for the agent's current state and observation to efficiently understand instruction.

\textbf{Step II. Navigation Scene Mapping}. Following IVLN-CE, to achieve agent's long-term planning in persistent large scenes, we use a \underline{S}cene \underline{M}apping \underline{M}odule (SMM) to create and save scene maps, including semantic map $\mathcal{M}^i_{t}[sem]$ and occupancy $\mathcal{M}^i_{t}[ocu]$ map. And these saved scene maps are cropped for partial crops $\mathcal{M}c^i_{t}[sem]$ and $\mathcal{M}c^i_{t}[ocu]$ based on the agent current pose, to achieve efficient encoding with a Map-Encoder $Encoder_{map}$ for map embeddings $\mathcal{Z}^m_{t}$:
\begin{equation}
\label{model1}
\mathcal{Z}^m_{t} = Encoder_{map}(\mathcal{M}c^i_{t}[sem],\mathcal{M}c^i_{t}[ocu]),
\end{equation}
where $Encoder_{map}(\cdot)$ consists of four CBRA blocks, each CBRA includes a convolution, a normalization, a ReLU activation, and an averaging pool, with a spatial attention for focused feature extraction, i.e., $Encoder_{map} = [Attn_{cot}({CBRA(\cdot)}) \times 4]$.
More details of the proposed Map-Encoder are available in our Supplementary Material.

\textbf{Step III. Low-Level Action Planner}: We propose an Exploration and Verification (EaV) strategy to further improve the robustness of sequential-horizon navigation. The EaV offers two distinct navigation modes: exploration mode proactively guides agents toward destination, while verification mode validate and corrects trajectory errors in time.

\subsection{High-Level Perception Planner} \label{IS}
SeqWalker uses a High-Level Perception Planner to understand user instructions. In the SH-VLN task, it is difficult for the navigation agent to logically understand and process complex long instructions, thus, we propose an \underline{I}nstruction \underline{S}egmentation \underline{M}odule (ISM) to segment global instructions to reduce the difficulty in understanding redundant instructions. The ISM decouples the sequential-horizon instruction into a series of sub-phrases, and the agent can focus on each sub-phrase sequentially for more accurate navigation. 

Different from prior VLN models that encode instructions globally, we encode local instructions for a more accurate implementation. Specifically, for each phrase $\mathcal{S}^{i}_k$ in instruction $\mathcal{I}^{i}$ is encoded ${\Psi}_{T}(S^{i}_k)$ by Text-Encoder (CLIP \cite{CLIP}), and the observed RGB image $\mathcal{R}^{i}_{t}$ is encoded ${\Psi}_{V}(\mathcal{R}^{i}_{t})$ by Vision-Encoder (CLIP). Then, the similarity $\theta^k_t$ between ${\Psi}_{T}(S^{i}_k)$ and ${\Psi}_{V}(\mathcal{R}^{i}_{t})$ is calculated, and the probability of each phrase $\mathcal{P}(\mathcal{S}^{i}_k)$ matching the current state is:
\begin{equation}
\label{model2}
\mathcal{P}(\mathcal{S}^{i}_k) = Softmax(\theta^k_t),
\end{equation}
\begin{equation}
\label{model3}
\theta^k_t = Sim({\Psi}_{T}(S^{i}_{k}),{\Psi}_{V}(\mathcal{R}^{i}_{t})),
\end{equation}
where $Sim(\cdot)$ denotes the cosine similarity of the two inputs. However, simply selecting the phrase with the maximum $\mathcal{P}(\mathcal{S}^{i}_k)$ may cause an incorrect similar selection. For more robust selection, we further calculate the entropy value

$\Phi^i_{t}$ for $\mathcal{P}(\mathcal{S}^{i}_k)$ to select the most suitable phrase $\mathcal{S}^{i}_{k^*}$:
\begin{equation}
\label{model4}
\Phi^i_{t} = \mathcal{H}(\mathcal{P}(\mathcal{S}^{i}_0), \mathcal{P}(\mathcal{S}^{i}_1), \dots, \mathcal{P}(\mathcal{S}^{i}_n)),
\end{equation}
where $\mathcal{H}(\cdot)$ denotes entropy calculation, and when $\Phi^i_{t}$ is less than threshold $\Phi_\lambda$, we take the phrase $\mathcal{S}^{n}_{k^*}$ with the maximum probability as the current focused local instruction:
\begin{equation}
\label{model5}
\mathnormal{k}^* = \arg\max_{k} \left( \mathcal{P}(\mathcal{S}^{i}_k)\right),
\end{equation}
otherwise, we take the global instruction $\mathcal{I}^{i}$ as the current navigation instruction to avoid incorrect instruction ignoring. And for richer semantic reasoning navigation, we use a lightweight LLM, i.e., Qwen-0.5b \cite{qwen}, as an Instruction-Encoder to obtain instruction embeddings $\mathcal{Z}^s_{t}$:
\begin{equation}
\label{model6}
\mathnormal{Z}^s_{t} = LLM(\alpha\cdot\mathcal{S}^{i}_{k^*} + (1-\alpha) \cdot \mathcal{I}^{i}),
\end{equation}
where $\alpha = 1$ if $\Phi^i_t < \Phi_\lambda$; otherwise, $\alpha = 0$. The LLM performs next token prediction for instruction encoding, and we take the embedding of the last hidden layer as the instruction embeddings $\mathcal{Z}^s_{t}$. This adaptive instruction encoding method facilitates agents in understanding complex instructions. 

\subsection{Low-Level Action Planner} \label{EaV}
SeqWalker uses a Low-Level Action Planner to provide specific navigation actions. In the SH-VLN task, the navigation multi-trajectories are long and sequentially strict. The agent may take an incorrect trajectory due to a biased understanding of the semantics of some instructions, and the current error often causes subsequent navigation to fail irreversibly. To improve the robustness of the navigation trajectory, we further propose an Exploration and Verification strategy as the Low-Level Action Planner for navigation.

\textbf{Exploration Navigation Mode}:
The Exploration Navigation Mode moves the agent toward the destination. The agent integrates the above embeddings in Exploration Mode for action prediction. Specifically, the obtained embeddings $\mathcal{Z}^{m}_{t}$, $\mathcal{Z}^{s}_{t}$ are jointly loaded into an \underline{A}ction \underline{O}utput \underline{H}ead (AOH) and are combined with the previous hidden state ${h}^i_{t-1}$ and action ${a}^i_{t-1}$ to predict the next action ${a}^i_{t}$:
\begin{equation}
\label{model9}
{a}^i_t, h^a_t= {AOH}(\mathcal{Z}^{m}_{t}, \mathcal{Z}^{s}_{t}, h^i_{t-1}, {a}^i_{t-1}),
\end{equation}
where ${AOH}(\cdot)$ consists of two Gated Recurrent Units (GRU) with cross attention blocks, referring \cite{IVLN}. The ${AOH}(\cdot)$ outputs the probability of each action $\mathcal{A}$, and we take the action corresponding to the maximum probability as ${a}^i_{t}$. ${a}^i_{t}$ moves agent toward destination.

\begin{algorithm}[t]
    \caption{Exploration and Verification Strategy}
    \renewcommand{\algorithmicrequire}{\textbf{Input:}}
    \label{alg:Al}
    \begin{algorithmic}[1]
        \REQUIRE Current observation $\mathcal{O}^i_t$, current action ${a}^i_t$, user's instruction $\mathcal{I}^i$, ver-threshold value $\delta_0$ and $\alpha$ from Eq. (\ref{model6}).
        
        \STATE Obtain the current RGB observation: $\mathcal{O}^i_t[R]$;
        \STATE Recall the last selected phrase: $S^{i}_{k^*_l}$, and compute the currently selected phrase or global instruction: $S^{i}_{k^*}$;
        \IF {$\alpha = 0$ or ${k^*}$ $=$ $({k^*_l}+1)$} 
        \STATE Predict the next action ${a}^i_t$ with Eq. (\ref{model9}).
        \ELSE 
        \STATE text\_embedding = ${\Psi}_{T}$($S^{i}_{k^*_l+1}$);
        \STATE rgb\_embedding = ${\Psi}_{V}$($\mathcal{O}^i_t[R]$);
        \STATE $\delta_t$ = $sim$(rgb\_embedding, text\_embedding).
        \IF {$\delta_t$ \ \textless \ $\delta_0$}  
        \STATE ${a}^i_t$ = \{``TURN\_BACK\_LAST\_STEP"\};
        \STATE $mode$ = {``Ver"}: mandatory setting of the next action to take the action corresponding to the second highest probability.
        \ELSE 
        \STATE Predict the next action ${a}^i_t$ with Eq. (\ref{model9}).
        \ENDIF
        \ENDIF
        \STATE Perform ${a}^i_t$ to obtain new observation state $\mathcal{O}^i_t$.
    \end{algorithmic}
\end{algorithm}

\textbf{Verification Navigation Mode}:
The Verification Navigation Mode corrects the errors on the current navigation trajectory. Based on Exploration Mode, the agent can actively search for destinations in navigation scenes. However, the agent unavoidably misjudges the direction of forks due to sequential-horizon navigation trajectories. We further propose a Verification Mode that utilizes the internal logical order of the instructions to correct navigation trajectories. Specifically, the order of the agent navigation process should match the sequence of the phrases. When performing the previous action $a^i_{t-1}$ and switching phrases, the new phrase $\mathcal{S}^{n}_{k^*}$ should be the next phrase of the last selected phrase $\mathcal{S}^{n}_{k^*_l}$, i.e., $\mathcal{S}^{n}_{k^*}=\mathcal{S}^{n}_{k^*_l+1}$. And during navigation, the agent's current observation $\mathcal{O}^i_t[R]$ should have a high similarity to the selected phrase $\mathcal{S}^{n}_{k^*_l+1}$. Based on this logic, we conduct two terms verification. Firstly, the agent determines whether the currently selected phrase is $\mathcal{S}^{n}_{k^*_l+1}$ (Term-I). Secondly, the agent compares the similarity $\delta_t$ between $\mathcal{S}^{n}_{k^*+1}$ and $\mathcal{O}^i_t[R]$ with a ver-threshold value $\delta_{0}$ (Term-II). Specifically, when the phrase selected $\mathcal{S}^{n}_{k^*}$ is not $\mathcal{S}^{n}_{k^*_l+1}$ and $\delta_{t} < \delta_{0}$, the agent returns to the previous position where the $a^i_{t-1}$ performed, where $\delta_{0}$ is a learnable parameter determined during navigation training for better adaptive navigation. And the specific steps of EaV strategy are summarized in Algorithm \ref{alg:Al}. 

\subsection{Learning Sequential-Horizon Navigation}

\subsubsection{\textbf{Sequential-Horizon Dataset Transformation.}}\label{SH} To evaluate the proposed SH-VLN task, we extend the IVLN IR2R-CE dataset \cite{IVLN} and propose a new benchmark. Two key aspects are considered in constructing this benchmark: how to make the single-stage navigation trajectories be stitched together into sequential-horizon trajectories, and how to generate long instructions with more discriminating details that closely align with navigation scenes. 

\textit{\textbf{a). Construction Sequential-Horizon Trajectories.}} Different from the one-to-one mapping between instruction and trajectory in IR2R-CE, the sequential-horizon navigation task involves a single instruction spanning multiple trajectories. To construct such trajectories, we select and connect pairs from the IR2R-CE dataset whose end- and start-points align. Corresponding instructions are concatenated to ensure semantic coherence using LLM, i.e., LLaMa-13B \cite{LLaMA}. The concatenation prompt template is:
\begin{leftbar}
\noindent
{{\small \textit{\textbf{User:}} \textit{\textbf{Please help logically connect two navigation instructions into one, ensuring semantic coherence, with the end of the first serving as the start of the second.}} \textless \textbf{INS1}\textgreater \textless \textbf{INS2}\textgreater} }
\end{leftbar} 

\begin{table*}[t]
    \centering
    \renewcommand{\arraystretch}{1}
    \setlength{\tabcolsep}{1mm} 
    \resizebox{\textwidth}{!}{ 
    \begin{tabular}{l ccccccc c ccccccc}
        \toprule
        
        & \multicolumn{7}{c}{Val-Seen}
        && \multicolumn{7}{c}{Val-Unseen} \\
        \cmidrule{2-8}
        \cmidrule{10-16}

        Comparisons
        & ~~~TL~~~
        & ~~OS$\uparrow$~~
        & nDTW$\uparrow$
        & ~SR$\uparrow$~
        & ~SPL$\uparrow$~
        & CPsubT$\uparrow$ 
        & t-nDTW$\uparrow$
        &
        & ~~~TL~~~
        & ~~OS$\uparrow$~~
        & nDTW$\uparrow$
        & ~SR$\uparrow$~
        & ~SPL$\uparrow$~
        & CPsubT$\uparrow$ 
        & t-nDTW$\uparrow$ \\
        
        \midrule
        
        CMA 
        & 12.5\tiny{$\pm$0.8} & 21\tiny{$\pm$1 } & 35\tiny{$\pm$3 } & 11\tiny{$\pm$3 } & 8\tiny{$\pm$2 } & 33\tiny{$\pm$1 } & 31\tiny{$\pm$2 }
        & & 12.1\tiny{$\pm$0.3} & 16\tiny{$\pm$1 } & 36\tiny{$\pm$2 } & 9\tiny{$\pm$2 } & 6\tiny{$\pm$1 } & 30\tiny{$\pm$2 } & 30\tiny{$\pm$1 } \\
        
        TourCMA 
        & 12.8\tiny{$\pm$0.6} & 23\tiny{$\pm$3 } & 36\tiny{$\pm$2 } & 14\tiny{$\pm$2 } & 10\tiny{$\pm$1 } & 38\tiny{$\pm$2 } & 32\tiny{$\pm$1 }
        & & 12.5\tiny{$\pm$0.3} & 19\tiny{$\pm$1 } & 33\tiny{$\pm$1 } & 11\tiny{$\pm$2 } & 8\tiny{$\pm$2 } & 38\tiny{$\pm$1 } & 28\tiny{$\pm$2 } \\
        
        PoolCMA 
        & 12.2\tiny{$\pm$0.7} & 19\tiny{$\pm$2 } & 33\tiny{$\pm$2 } & 11\tiny{$\pm$3 } & 9\tiny{$\pm$2 } & 36\tiny{$\pm$1 } & 29\tiny{$\pm$2 }
        & & 12.3\tiny{$\pm$0.6} & 18\tiny{$\pm$1 } & 31\tiny{$\pm$1 } & 13\tiny{$\pm$1 } & 10\tiny{$\pm$2 } & 34\tiny{$\pm$2 } & 27\tiny{$\pm$2 } \\
        
        Po.E.CMA 
        & 12.2\tiny{$\pm$0.8} & 22\tiny{$\pm$3 } & 37\tiny{$\pm$3 } & 13\tiny{$\pm$2 } & 11\tiny{$\pm$2 } & 42\tiny{$\pm$1 } & 33\tiny{$\pm$2 }
        & & 11.6\tiny{$\pm$0.7} & 20\tiny{$\pm$3 } & 38\tiny{$\pm$2 } & 12\tiny{$\pm$2 } & 10\tiny{$\pm$1 } & 41\tiny{$\pm$3 } & 32\tiny{$\pm$2 } \\
        
        MAP-CMA 
        & 14.5\tiny{$\pm$0.6} & 35\tiny{$\pm$3} & 46\tiny{$\pm$3} & 26\tiny{$\pm$3} & 22\tiny{$\pm$2} & 53\tiny{$\pm$1 } & 43\tiny{$\pm$1}
        & & 13.5\tiny{$\pm$0.6} & 31\tiny{$\pm$2} & 44\tiny{$\pm$2} & 23\tiny{$\pm$1} & 20\tiny{$\pm$1} & 50\tiny{$\pm$1 } & 39\tiny{$\pm$1} \\
        
        ETPNav  
        & 13.9\tiny{$\pm$0.7} & 30\tiny{$\pm$3} & 41\tiny{$\pm$2} & 23\tiny{$\pm$2} & 20\tiny{$\pm$1} & 46\tiny{$\pm$2 } & 38\tiny{$\pm$1}
        & & 13.1\tiny{$\pm$0.5} & 29\tiny{$\pm$2} & 38\tiny{$\pm$1} & 20\tiny{$\pm$1} & 16\tiny{$\pm$1} & 44\tiny{$\pm$1 } & 32\tiny{$\pm$1} \\
        
        HNR 
        & 14.0\tiny{$\pm$0.6} & 31\tiny{$\pm$3} & 42\tiny{$\pm$3} & 24\tiny{$\pm$4} & 21\tiny{$\pm$2} & 49\tiny{$\pm$1 } & 39\tiny{$\pm$1}
        & & 13.3\tiny{$\pm$0.7} & 30\tiny{$\pm$2} & 39\tiny{$\pm$2} & 21\tiny{$\pm$1} & 17\tiny{$\pm$1} & 45\tiny{$\pm$2 } & 33\tiny{$\pm$2} \\
        
        \midrule
        SeqWalker
        & 17.8\tiny{$\pm$1.2} & \bf 43\tiny{$\pm$2 } & \bf 52\tiny{$\pm$2 } & \bf 34\tiny{$\pm$1 } & \bf 33\tiny{$\pm$2 } & \bf 67\tiny{$\pm$2 } & \bf 48\tiny{$\pm$1 }
        & & 17.3\tiny{$\pm$1.0} & \bf 40\tiny{$\pm$2 } & \bf 50\tiny{$\pm$1 } & \bf 30\tiny{$\pm$2 } & \bf 29\tiny{$\pm$1 } & \bf 66\tiny{$\pm$3 } & \bf 45\tiny{$\pm$1 } \\
        
        \bottomrule
    \end{tabular}
    } 
    \caption{The test results for SeqWalker compared to SOTA methods on the \textbf{\textit{SH IR2R-CE datasets}}. TL is in meters, and OS, nDTW, SR, SPL, CPsubT, t-nDTW are reported as percentages. Results are presented as mean ± standard deviation.}
    \label{TAB1}
    \vspace{-2mm}
\end{table*}

\begin{table*}[t]
    \centering
    \renewcommand{\arraystretch}{1}
    \setlength{\tabcolsep}{1.1mm} 
    \resizebox{\textwidth}{!}{ 
    \begin{tabular}{l ccccccc c ccccccc}
        \toprule
        & \multicolumn{7}{c}{Val-Seen}
        && \multicolumn{7}{c}{Val-Unseen} \\
        \cmidrule{2-8}
        \cmidrule{10-16}

        Comparisons
        & ~~~~~TL~~~
        & ~~~NE$\downarrow$~~~
        & ~~OS$\uparrow$~~
        & nDTW$\uparrow$
        & ~SR$\uparrow$~
        & ~SPL$\uparrow$~
        & t-nDTW$\uparrow$
        &
        & ~~~~~TL~~~
        & ~~~NE$\downarrow$~~~
        & ~~OS$\uparrow$~~
        & nDTW$\uparrow$
        & ~SR$\uparrow$~
        & ~SPL$\uparrow$~
        & t-nDTW$\uparrow$ \\
        \midrule

        CMA 
        & 7.8\tiny{$\pm$0.4} & 8.8\tiny{$\pm$0.6} & 27\tiny{$\pm$3} & 42\tiny{$\pm$3} & 18\tiny{$\pm$3} & 17\tiny{$\pm$3} & 39\tiny{$\pm$1}
        & ~
        & 7.5\tiny{$\pm$0.3} & 8.8\tiny{$\pm$0.2} & 26\tiny{$\pm$1} & 44\tiny{$\pm$1} & 19\tiny{$\pm$1} & 18\tiny{$\pm$1} & 38\tiny{$\pm$2} \\

        TourCMA 
        & 8.0\tiny{$\pm$0.4} & 8.2\tiny{$\pm$0.9} & 30\tiny{$\pm$2} & 44\tiny{$\pm$2} & 20\tiny{$\pm$3} & 19\tiny{$\pm$2} & 40\tiny{$\pm$1}
        & ~
        & 7.8\tiny{$\pm$0.1} & 9.0\tiny{$\pm$0.2} & 26\tiny{$\pm$1} & 42\tiny{$\pm$1} & 18\tiny{$\pm$0} & 17\tiny{$\pm$1} & 36\tiny{$\pm$1} \\

        PoolCMA 
        & 7.2\tiny{$\pm$0.5} & 9.1\tiny{$\pm$0.4} & 24\tiny{$\pm$4} & 41\tiny{$\pm$2} & 17\tiny{$\pm$4} & 16\tiny{$\pm$2} & 37\tiny{$\pm$2}
        &
        & 7.3\tiny{$\pm$0.2} & 9.0\tiny{$\pm$0.3} & 23\tiny{$\pm$1} & 42\tiny{$\pm$1} & 16\tiny{$\pm$1} & 15\tiny{$\pm$0} & 36\tiny{$\pm$2} \\

        Po.E.CMA 
        & 7.6\tiny{$\pm$0.8} & 8.9\tiny{$\pm$0.9} & 27\tiny{$\pm$3} & 42\tiny{$\pm$3} & 18\tiny{$\pm$4} & 17\tiny{$\pm$2} & 38\tiny{$\pm$2}
        & ~
        & 6.9\tiny{$\pm$0.2} & 8.7\tiny{$\pm$0.2} & 25\tiny{$\pm$2} & 44\tiny{$\pm$1} & 18\tiny{$\pm$1} & 16\tiny{$\pm$1} & 38\tiny{$\pm$2} \\

        MAP-CMA 
        & 9.4 & 6.4 & 48 & 56 & 39 & 36 & 52
        & ~
        & 8.5 & 6.8 & 44 & 54 & 35 & 32 & 47 \\

        ETPNav
        & 9.6\tiny{$\pm$0.9} & 8.8\tiny{$\pm$0.9} & 45\tiny{$\pm$4} & 52\tiny{$\pm$2} & 36\tiny{$\pm$2} & 31\tiny{$\pm$1} & 47\tiny{$\pm$2}
        & ~
        & 8.9\tiny{$\pm$0.6} & 8.6\tiny{$\pm$0.2} & 37\tiny{$\pm$2} & 47\tiny{$\pm$2} & 28\tiny{$\pm$1} & 27\tiny{$\pm$1} & 41\tiny{$\pm$2} \\

        HNR
        & 10.8\tiny{$\pm$0.5} & 7.9\tiny{$\pm$0.9} & 46\tiny{$\pm$4} & 53\tiny{$\pm$3} & 37\tiny{$\pm$2} & 33\tiny{$\pm$2} & 50\tiny{$\pm$1}
        & ~
        & 9.3\tiny{$\pm$0.7} & 7.5\tiny{$\pm$0.4} & 40\tiny{$\pm$2} & 50\tiny{$\pm$1} & 30\tiny{$\pm$1} & 28\tiny{$\pm$2} & 44\tiny{$\pm$2} \\
        
        OVER-NAV 
        & 9.5\tiny{$\pm$0.9} & 5.8\tiny{$\pm$0.9} & 49\tiny{$\pm$4} & 59\tiny{$\pm$2} & 39\tiny{$\pm$2} & 36\tiny{$\pm$2} & 56\tiny{$\pm$2}
        & ~
        & 8.8\tiny{$\pm$0.6} & 6.5\tiny{$\pm$0.2} & 45\tiny{$\pm$2} & 56\tiny{$\pm$1} & 35\tiny{$\pm$1} & 33\tiny{$\pm$1} & 50\tiny{$\pm$2} \\

        \midrule
        SeqWalker
        & 12.3\tiny{$\pm$0.8} & \bf 5.5\tiny{$\pm$0.7} & \bf 51\tiny{$\pm$3} & \bf 60\tiny{$\pm$2} & \bf 41\tiny{$\pm$2} & \bf 38\tiny{$\pm$2} & \bf 58\tiny{$\pm$2}
        & ~
        & 11.4\tiny{$\pm$0.5} & \bf 6.4\tiny{$\pm$0.2} & \bf 46\tiny{$\pm$2} & \bf 58\tiny{$\pm$2} & \bf 36\tiny{$\pm$2} & \bf 34\tiny{$\pm$2} & \bf 52\tiny{$\pm$2} \\
        
        \bottomrule
    \end{tabular}
    } 
    \caption{
    The test results for SeqWalker compared to SOTA methods on the \textbf{\textit{IR2R-CE datasets}} \cite{IVLN}. TL and NE are in metres, and OS, nDTW, SR, SPL, and t-nDTW are reported as percentages. The results are reported as $\bar{x} \pm \sigma_{\bar{x}}$.
    }
    \label{TAB2}
    \vspace{-3mm}
\end{table*}

\begin{figure}[t!]
\centering
\includegraphics[width=0.95\linewidth]{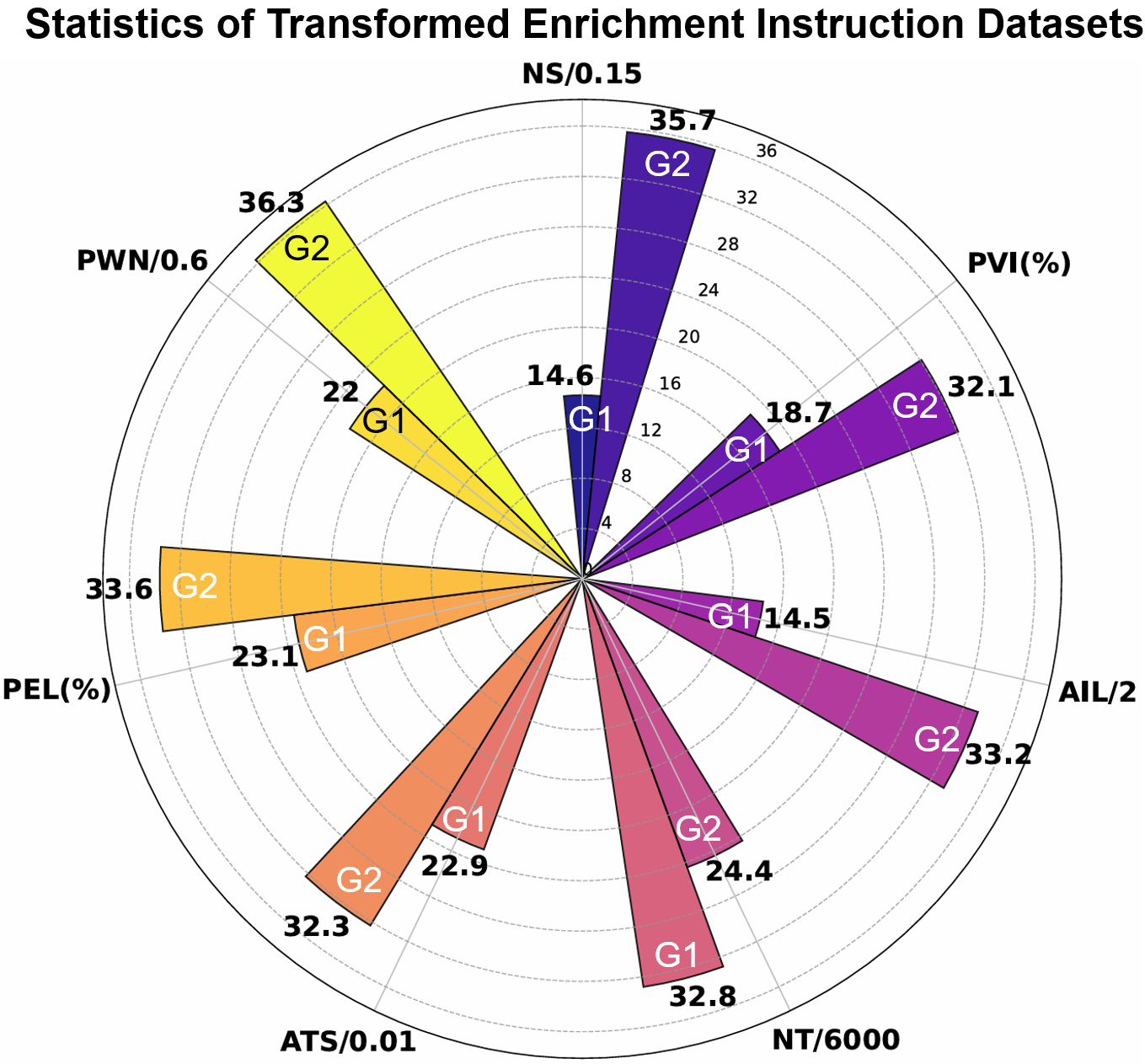}
\caption{Statistics of the transformed datasets. The statistics graph contains seven groups of comparison metrics: \underline{N}umber of \underline{S}entences, 
\underline{P}ercentage of \underline{V}erbs in \underline{I}nstruction, \underline{A}verage \underline{I}nstruction \underline{L}ength, \underline{N}umber of \underline{T}rajectories, \underline{P}ercentage of \underline{E}xplicit \underline{L}ocations, \underline{A}verage \underline{T}rajectories \underline{S}cene, \underline{P}hrases \underline{W}ord \underline{N}umber, \textbf{G1} represents unchanged instructions, and \textbf{G2} represents our transformed enrichment IR2R-CE dataset. Please note that for better visualization, we mark some metrics as ``$A/x$", i.e., and the true value is $A \times x$. 
}
\vspace{-3mm}
\label{SSF_3}
\end{figure}

\textit{\textbf{b). Enrichment Long Instructions.}} In addition to generating sequential-horizon trajectories, we also enrich the instructions to address their lack of sufficient discriminative granularity to distinguish semantically similar subtasks in multi-trajectory navigation. Specifically, we extend the IVLN instructions based on a large multimodal language model, i.e., LLaVA-OneVision \cite{LLaVAQwen}, that can reason multiview images with language prompts. Specifically, during instruction enrichment, an agent is forced to follow the ground-truth trajectories for step-by-step navigation. In the navigation trajectory corresponding to each phrase $S^{i}_k$, we collect the observed RGB images to construct multiple images set $\{R^{k}_0, R^{k}_1, \dots, R^{k}_n\}$ of a first view corresponding to a navigation trajectory. Then, we feed the phrase $S^{i}_k$ and multiple images as inputs into LLaVA to implement instruction enrichment. The used enrichment prompt template is: 

\begin{leftbar}
\noindent
{{\small \textit{\textbf{User:}} \textless \textbf{IMAGES}\textgreater \ \textit{\textbf{Please look closely at these multiple images of the first view corresponding to a navigation trajectory, and help me enrich the discriminating details of instruction without changing its logic. The original instruction is:}} \textless \textbf{INS}\textgreater \ \textbf{Once you enter the bed room, go straight to the left-hand door near the table.}}}
\end{leftbar} 

\noindent where the bold is the fixed enrichment prompt and the \textless \textbf{INS}\textgreater \ part is the user instruction phrase. The \textless{\textbf{IMAGES}}\textgreater \ represents the tokens containing the multiple images set $\{R^{k}_0, R^{k}_1, \dots, R^{k}_n\}$ of a first view corresponding to a navigation trajectory. We save the enriched instruction output provided by LLaVA and replace phrases $S^{i}_k$. The process is repeated to perform the expansion for all instructions.

Based on the above two main aspects, we construct the \underline{S}equential-\underline{H}orizon IR2R-CE (SH IR2R-CE), and a large-scale collection of single trajectories paired with corresponding instructions is used for training and testing. The statistics for the constructed dataset are presented in Fig.\ref{SSF_3}

\section{Experiments}

\subsection{Implementation}
\vspace{-0.5mm}
We use Habitat \cite{Habitat} as the simulation platform. Our SeqWalker is trained on the proposed SH IR2R-CE dataset. We use a lightweight LLM, i.e., Qwen-0.5b \cite{qwen}, as an Instruction-Encoder to obtain instruction embeddings, and CLIP ViT-B/32  as the image-text encoders for the high-level perception planner, and we set the threshold $\Phi_\lambda = 0.65$, the Term-I and Term-II for turning back are included in verification mode. We evaluate the performance following previous metrics \cite{M1, M2, IVLN}: Navigation Error (NE), Oracle Success Rate (OS), normalized Dynamic Time Warping (nDTW), Success Rate (SR), Success weighted by Path Length (SPL), Tour normalized Dynamic Time Warping (t-nDTW), and Trajectory Length (TL). We also propose two metrics to evaluate sub-instruction task completion: $CP_{sub}T$, the ratio of completed sub-tasks before failure, $CP_{sub}T = NS / NA$, where $NS$ and $NA$ are the numbers of completed and total sub-tasks, respectively. $CP_{sub}I$, the proportion of correctly selected sub-instructions, defined as $CP_{sub}I = NC / NT$, where $NC$ is the number of correct selections based on entropy and $NT$ is the total steps.

\subsubsection{\textbf{Training Strategy.}} The proposed SeqWalker adopts a two-stage imitation training strategy, following \cite{VLNCE}. The first stage of the training process uses teacher forcing on the proposed SH-IR2R-CE datasets. In the second stage, the model of the first stage is fine-tuned to achieve a better generalization performance in unseen scenes. We use the Progress Monitor auxiliary loss in both training steps. The modules in SeqWalker are partially trainable, as shown in Fig.\ref{fig_2}. We freeze the parameters of CLIP models to preserve their text-image matching and understanding capabilities. The LLM-based Instruction-Encoder, Maps-Encoder, and AOH are jointly trained during the training.

\subsection{Comparison Experiments}

\vspace{-0.5mm}
\subsubsection{\textbf{How does SeqWalker perform with sequential-horizon instructions?}} 
This experiment is used to verify the performance of SeqWalker on the sequential-horizon navigation task. To ensure fairness, all methods are trained on the proposed SH IR2R-CE dataset. The SOTA comparison models include CMA \cite{IVLN} series are Naive-CMA, PoolCMA, PoolEndCMA, etc. and HNR \cite{VLN241}, ETPNav \cite{ETP}. A summary of the evaluation results can be found in Table~\ref{TAB1}. SeqWalker achieves superior navigation performance with long instructions, improving the t-nDTW metric by 5\% and 6\% over the previous best model on the val-seen and val-unseen sets, respectively. Significant improvements are observed across multiple metrics, e.g., SR, t-nDTW, etc. For methods without local instruction awareness, $CP_{sub}I$ is 0, whereas SeqWalker achieves 74\%, demonstrating effective sub-instruction comprehension during navigation. The results show that SeqWalker has superior performance in the SH-VLN task.

\subsubsection{\textbf{How does SeqWalker perform with traditional single-task instructions?}} We also conduct experiments to verify the navigation performance on the traditional IVLN task. The SOTA comparison models also include the CMA family \cite{IVLN} of Naive-CMA, PoolCMA, PoolEndCMA, MAP-CMA, and OVER-NAV \cite{OverVLN}. A summary of the results can be found in Table~\ref{TAB2}. Leveraging the effectiveness of the hierarchical planning framework, SeqWalker also has superior performance on the IVLN task.

\begin{table}[t]
\centering
\setlength{\tabcolsep}{0.6pt}  
\renewcommand{\arraystretch}{1}  
\normalsize                
\begin{tabular}{c c c | c c c c c c c}
\toprule
\# & ISM & LLM-Size & TL & NE$\downarrow$ & OS$\uparrow$ & nDTW$\uparrow$ & SR$\uparrow$ & SPL$\uparrow$ & t-nDTW$\uparrow$ \\
\midrule
1 & \cmark     & 0.5B  & 17.3 & \textbf{6.6} & \textbf{40} & \textbf{50} & \textbf{30} & \textbf{29} & \textbf{45} \\
\midrule 
2 & \xmarkg     & 0.5B  & 17.5 & 7.9 & 33 & 44 & 25 & 21 & 38 \\
3 & \xmarkg     & 7.0B  & 17.2 & 7.3 & 36 & 46 & 27 & 23 & 41 \\
4 & \xmarkg    & 13B   & 16.8 & 7.0 & 37 & 48 & 27 & 24 & 42 \\
\bottomrule
\end{tabular}
\vspace{-2mm}   
 \caption{Ablation study on Val-Unseen split. Instruction encoding consistently improves with larger LLMs, while ISM brings notable gains for smaller and mid-sized ones.}
\label{tab:ablation_full}
\end{table}

\begin{table}[t]
    \centering
    \renewcommand{\arraystretch}{1}
    \setlength{\tabcolsep}{1.1pt} 
    \begin{tabular}{l|ccccccc}
        \toprule
        \small{$\#$} ~Model 
        & TL 
        & NE$\downarrow$ 
        & OS$\uparrow$ 
        & nDTW$\uparrow$ 
        & SR$\uparrow$ 
        & SPL$\uparrow$ 
        & t-nDTW$\uparrow$ \\
        \midrule
        1 ~$Maximum$ 
        & 17.2 & 7.3 & 36 & 47 & 27 & 25 & 42 \\
        \midrule
        2 ~$\Phi_\lambda = 0.85$ 
        & 17.5 & 7.2 & 38 & 48 & 27 & 26 & 43 \\
        3 ~$\Phi_\lambda = 0.75$ 
        & 16.8 & 6.9 & 39 & 48 & 29 & 28 & 44 \\
        4 ~$\Phi_\lambda = 0.65$ 
        & 17.3 & \bf 6.6 & \bf 40 & \bf 50 & \bf 30 & \bf 29 & \bf 45 \\
        5 ~$\Phi_\lambda = 0.55$ 
        & 17.4 & 7.0 & 37 & 47 & 28 & 26 & 43 \\
        6 ~$\Phi_\lambda = 0.45$ 
        & 17.2 & 7.4 & 37 & 46 & 27 & 25 & 42 \\
        \bottomrule
    \end{tabular}
    \caption{Ablation study for \textbf{phrase selection method} and different \textbf{\mbox{entropy threshold values} $\Phi_\lambda$} on the Val-Unseen.}
    \label{TAB4}
    \vspace{-3mm}
\end{table}

\begin{table}[t]
    \centering
    \renewcommand{\arraystretch}{0.9}
    \setlength{\tabcolsep}{2.3pt}  
    \setlength{\tabcolsep}{0.61mm} 
    \begin{tabular}{l|ccccccc}
        \toprule
        \small{$\#$} ~Model 
        & TL 
        & NE$\downarrow$ 
        & OS$\uparrow$ 
        & nDTW$\uparrow$ 
        & SR$\uparrow$ 
        & SPL$\uparrow$ 
        & t-nDTW$\uparrow$ \\
        \midrule
        1 ~Type-I
        & 16.4 & 8.2 & 32 & 45 & 23 & 21 & 39 \\
        
        2 ~Type-II
        & 16.7 & 6.9 & 36 & 46 & 26 & 24 & 42 \\
        
        3 ~Type-III
        & 16.7 & 7.1 & 36 & 45 & 25 & 24 & 41 \\
        
        \midrule
        4 ~Type-IV
        & 17.3 & \bf 6.6 & \bf 40 & \bf 50 & \bf 30 & \bf 29 & \bf 45 \\
        \bottomrule
    \end{tabular}
    \caption{\mbox{Ablation study of the proposed \textbf{ISM} on Val-Unseen.}}
    \label{TAB3}
    \vspace{-2mm}
\end{table}

\subsection{Ablation Studies}

\subsubsection{\textbf{What are the advantages of our ISM over simply using LLM with a larger number of parameters?}} We study the effectiveness of ISM improving instruction encoding, and Table \ref{tab:ablation_full} summarizes the ablation results. Although instruction encoding improves with increasing LLM parameters, our ISM significantly enhances LLM performance with smaller models. Our ISM splits long instructions into sub-tasks, which significantly improves the reasoning of small-parameter LLM. This is crucial in real-world robotics applications, where real-time constraints and limited on-device computation hinder the deployment of large LLMs.

\subsubsection{\textbf{What is the best entropy threshold for segmentation?}}We first provide experiment results for simply selecting phrases with maximum similarity, denoted as $Maximum$. We also study the effect of different entropy thresholds $\Phi_\lambda$ on navigation performance; ablation results are summarized in Table~\ref{TAB4}. As $\Phi_\lambda$ decreases, navigation performance initially improves, peaking near $\Phi_\lambda = 0.65$, then decreases (see Table~\ref{TAB4} “\#2-6”). Higher thresholds allow more phrases to be used, improving navigation. Excessive $\Phi_\lambda$ harms prediction from global instructions and causes incorrect choices among similar phrases, reducing performance.

\subsubsection{\textbf{How do different segmentation styles perform?}} We study the effectiveness of instruction segmentation and the effect of different segmentation styles on navigation performance, and the ablation results are summarized in Table \ref{TAB3}. Type-I indicates the instructions are not segmented. Type-II indicates that the instructions are segmented based on commas. Type-III indicates that the instructions are segmented using conjunctions such as ``and" or ``then". Type-IV indicates that the instructions are segmented based on period. The results demonstrate that instruction segmentation has a significant improvement on sequential-horizon instruction navigation (referring Table \ref{TAB3} ``\#1"). The different segmentation styles  have a significant affects navigation performance. Type-IV achieves the best results, as its comma-based segmentation ensures each phrase includes at least one object, facilitating better alignment with visual observations.

\subsubsection{\textbf{How effective is the EaV strategy?}} We study the effect of the proposed EaV strategy, including Term-I and Term-II in Verification Mode. The ablation results are summarized in Table \ref{TAB6}. This ablation study demonstrates that the proposed EaV strategy can significantly improve the success rate of navigation. Although the trajectory correction procedures introduce additional trajectory length, they contribute to notable improvements in navigation success rates and other related metrics, e.g., task completion and trajectory quality. When Term-I and Term-II are both included in EaV strategy, our SeqWalker achieves robust sequential-horizon navigation with the best performance (referring Table \ref{TAB6} ``\#4").

\begin{table}[t]
\centering
    \renewcommand{\arraystretch}{0.9}
    \setlength{\tabcolsep}{0.01mm} 
	\begin{tabular}{cc c c| ccccccc c ccccccc}
			\toprule

			\small{\#} & \small{Ver.}  & \small{Term-I} &  \small{Term-II} 
                
                & TL
			&  NE$\downarrow$
			&  OS$\uparrow$
			&  nDTW$\uparrow$
			&  SR$\uparrow$
			&  SPL$\uparrow$
			&  t-nDTW$\uparrow$
			\\
			\midrule
			1
                & \xmarkg 
                & \xmarkg
                & \xmarkg
                & 12.9
                & 7.4
                & 35
                & 44
                & 25
                & 23
                & 40
			\\
                \midrule
                2
                & \cmark
                & \cmark 
                & \xmarkg
                & 17.8
                & 6.8
                & 38
                & 48
                & 28
                & 27
                & 43
			\\
                3
                & \cmark
                & \xmarkg
                & \cmark
                & 17.1
                & 7.1
                & 38
                & 47
                & 27
                & 25
                & 42
			\\
                4
                & \cmark
                & \cmark 
                & \cmark
                &  17.3
                & \bf 6.6
                & \bf 40
                & \bf 50
                & \bf 30 
                & \bf 29
                & \bf 45
			\\
			\bottomrule
	\end{tabular}
	\caption{The ablation study for the proposed \textbf{EaV strategy} on the proposed SH IR2R Val-Unseen Scenes.}
	\label{TAB6}
\vspace{-4mm}
\end{table}

\section{Conclusion}
In this paper, we introduce a new VLN task called Sequential-Horizon Vision-and-Language Navigation. To solve the SH-NLV task, we propose SeqWalker, a novel navigation model built on a hierarchical planning framework. In High-Level Planner, SeqWalker performs local segmentation on global long instructions to obtain the most relevant segmented instruction with the agent's current scenes. In Low-Level Planner, we develop an Exploration and Verification strategy to achieve forward progress toward the destination or to leverage the inherent logical order of instructions to dynamically correct the error of current navigation trajectories. To evaluate the SH-VLN performance, we extend the IVLN dataset and establish a new benchmark. Extensive experiments validate SeqWalker's superior performance.

\newpage

\section{Acknowledgments}
This work was supported by the National Key Research and Development Program of China under Grant 2024YFB4707700, the National Natural Science Foundation of China under Grant T2596045, T2596040 and U23A20343, CAS Project for Young Scientists in Basic Research, Grant YSBR-041, Liaoning Provincial ``Selecting the Best Candidates by Opening Competition Mechanism" Science and Technology Program under Grant 2023JH1/10400045, Fundamental Research Project of SIA under Grant 2024JC3K01.

\bibliography{aaai2026}

\section{Notes on Supplementary Material}

This supplementary material provides additional details regarding the proposed methods, experimental setups, and results. A summary of the contents is outlined below:
\begin{itemize}

    \item The details of the comparison methods and the experiment implementation are provided in Section \ref{experiment}.

    \item The evaluation metrics details are provided in Section \ref{METRICSR}.

    \item The more details on ablation studies and methodological efficiency comparisons in Section \ref{AS}.

    \item Section \ref{entropyT} presents a detailed explanation of the method used to determine the entropy threshold $\Phi_\lambda$ for ISM.  

    \item Ablation studies and discussions on verification threshold (in Verification Mode) are provided in Section \ref{VER}.  
     
    \item The details of Map-Encoder are provided in Section \ref{MAPEENCODER}.

    \item The statistical details and more examples of the proposed SH IR2R-CE dataset are provided in Section \ref{LSLI-IR2R-CE}.
    
    \item The implementation details of different instruction segmentation styles are provided in Section \ref{STYLES}.

     \item More visualization examples are provided in Section \ref{VVLN}.
    
    \item Discussions and future work are provided in Section \ref{DaFW}.

\end{itemize}

\section{Comparison Methods and Implementation}\label{experiment}

We use Habitat \cite{Habitat} as the simulation platform for our experiments. SeqWalker is trained on the proposed SH IR2R-CE dataset mentioned in section 3.5. We use CLIP ViT-B/32 \cite{CLIP} as the image-text encoders, and we set threshold $\Phi_\lambda = 0.65$, all the terms of turning back are included in verification mode. About dataset construction, we load LLaVA-OneVision-qwen2-7b \cite{LLaVAQwen} to expand the original instructions. We also leverage LLaMA-3-8B \cite{LLaMA} for the construction of long trajectories. The experiments are conducted using PyTorch 2.1.2 (CUDA 12.4) with eight NVIDIA RTX 3090 GPUs. One GPU is allocated for model computing, while the other seven are for environment simulation.

\textbf{SH IR2R-CE Dataset Comparison:}
The proposed benchmark SH IR2R-CE dataset is used for training and evaluation. Our comparison methods include \textbf{no scene memory} methods: Cross-Modal Attention (CMA) \cite{VLNCE}, which is a classical baseline in continuous environment VLN and without cross-episode reasoning ability; the HNR \cite{VLN241} and ETPNav \cite{ETP} are the latest VLN-CE models available for R2R-CE settings, without cross-episode reasoning ability. The \textbf{scene memory} methods include: TourCMA \cite{IVLN}, which resets its hidden state of GRU only at the start of each tour, extending the receptive field to all tours; PoolCMA \cite{IVLN}, which resets its hidden state at the beginning of each episode, but temporarily max-pooling the hidden state into a tour-persistent vector is performed at the end of each tour; PoolEndCMA \cite{IVLN}, which combines the PoolCMA tour memory with the final state vector, then uses the resulting data to predict the next action. To ensure fairness, we train all the comparison methods in the Implicit IR2R-CE dataset. Following the previous settings \cite{OverVLN,IVLN}, we run each experiment 3 times and report mean and variance metrics as $\bar{x} \pm \sigma_{\bar{x}}$. 

\noindent \textbf{IR2R-CE Dataset Comparison:}
The common public benchmark IR2R-CE dataset \cite{IVLN} is used for training and evaluation. Our comparison methods include: Cross-Modal Attention (CMA) \cite{VLNCE}, TourCMA \cite{IVLN}, PoolCMA \cite{IVLN}, PoolEndCMA \cite{IVLN}, HNR \cite{VLN241}, ETPNav \cite{ETP} and OVER-NAV \cite{OverVLN}, which is an IVLN model with open-vocabulary detection and structured representation. The results of all comparison methods are reported in previous methods \cite{OverVLN,VLNCE}. Following the previous settings \cite{OverVLN,IVLN}, we run each experiment 3 times and report mean and variance metrics as $\bar{x} \pm \sigma_{\bar{x}}$. 

\section{Evaluation Metrics}\label{METRICSR}
Following previous studies\cite{VLNCE,IVLN}, we use comprehensive evaluation metrics. They cover various aspects such as navigation precision, efficiency, and adherence to instructions. In addition, we also provide two new metrics to accurately evaluate the completion of each subtask: $CPsubT$ and $CPsubI$.

\noindent\textbf{\underline{C}om\underline{p}leted \underline{sub}-\underline{T}asks Ratio (CPsubT)} is defined as the percentage of sub-tasks completed before failure:
\[
\mathrm{CPsubT} = \frac{N_S}{N_A},
\]
where $N_S$ is the number of successfully completed sub-tasks, and $N_A$ is the total number of assigned sub-tasks. A value of $\mathrm{CPsubT} = 1$ indicates that all sub-tasks are completed successfully, whereas $\mathrm{CPsubT} = 0$ means all sub-tasks failed.

\noindent\textbf{Correct Percentage of sub-Instruction (CPsubI)} is defined as the percentage of correct sub-instruction selections during navigation: $\mathrm{CPsubI} = \frac{N_C}{N_T}$, where $N_C$ is the number of steps at which the correct sub-instruction is selected (based on entropy), and $N_T$ is the total number of navigation steps. Selecting a global instruction instead of the appropriate sub-instruction is counted as an incorrect decision. Note that due to the comparison methods not having local instruction awareness, the calculated CPsubI is 0. SeqWalker achieves 74\%.

\begin{figure}[t!]
\centering
\includegraphics[width=1\linewidth]{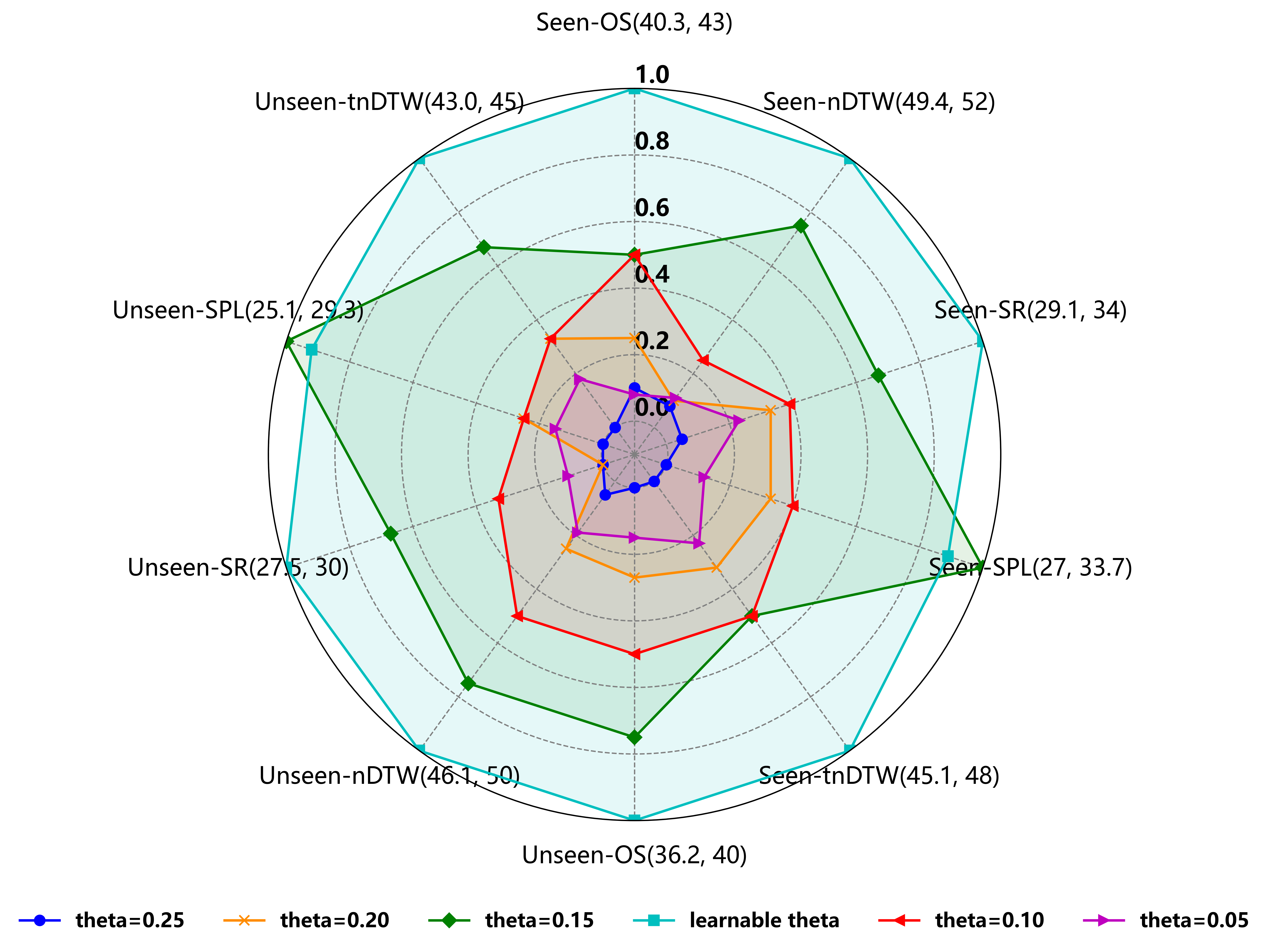}
\caption{The ablation studies of different \textbf{verification thresholds} on the SH IR2R-CE dataset. For fixed thresholds, it starts at 0.05 and gradually increases to 0.25, with a pace of 0.05. The learnable thresholds $\delta_{0}$ achieve the best performance. Note that for better visualization, we normalize the evaluation metrics and label the original values ($Maximum$, $Minimum$).}
\label{fig_4}
\vspace{-3mm}
\end{figure}

\noindent\textbf{\underline{T}rajectory \underline{L}ength (TL)}: The TL represents the total length of the navigation trajectory from the starting position to the terminal position. It is the sum of the distances between all adjacent nodes along the path. Let \( T_0 \) be the starting point and \( T_n \) be the endpoint, then the path length is given by:
\begin{equation}
\label{model2}
TL = \sum_{i=0}^{n-1} d(T_i, T_{i+1}),
\end{equation}
where \( d(T_i, T_{i+1}) \) is the distance between two adjacent nodes along the path.

\noindent\textbf{\underline{N}avigation \underline{E}rror (NE)}:
The navigation error refers to the distance between the endpoint of the predicted path and the endpoint of the reference path (ground truth-path). Let \( T_n^{pred} \) be the endpoint of the predicted path and \( T_n^{ref} \) be the endpoint of the reference path, the navigation error is given by:
\begin{equation}
\label{model3}
NE = d(T_n^{pred}, T_n^{ref}).
\end{equation}

\noindent\textbf{Navigation \underline{S}uccess \underline{R}ate (SR)}: The navigation success rate indicates whether the distance between the predicted and reference path endpoints is within a predefined threshold \( \epsilon \). The formula is:
\begin{equation}
\label{model4}
SR = 
\begin{cases} 
1, & \text{if } d(T_n^{pred}, T_n^{ref}) \leq \epsilon \\
0, & \text{if } d(T_n^{pred}, T_n^{ref}) > \epsilon
\end{cases},
\end{equation}
where \( \epsilon \) is the predefined success distance threshold.

\noindent\textbf{\underline{O}racle \underline{S}uccess Rate (OS)}:
The Oracle Success Rate evaluates whether the distance from any point on the path to the goal point is within a predefined threshold. If the minimum distance from any node on the path to the goal point is less than or equal to the threshold, it returns 1; otherwise, it returns 0. The formula is:
\begin{equation}
\label{model5}
OS = 
\begin{cases} 
1, & \text{if } \min_{i \in [1, n]} d(T_i, T_t) \leq \epsilon \\
0, & \text{if } \min_{i \in [1, n]} d(T_i, T_t) > \epsilon
\end{cases},
\end{equation}
where \( T_t \) is the goal point, \( d(T_i, T_t) \) is the distance from node \( T_i \) on the path to the goal point, and \( \epsilon \) is the threshold distance.

\noindent\textbf{\underline{S}uccess Rate Weighted by \underline{P}ath \underline{L}ength (SPL)}: SPL considers both the success rate and the path length, penalizing overly long (inefficient) paths. The formula is:
\begin{equation}
\label{model2}
SPL = \frac{\text{Success Rate} \times TL}{TL_{ref}},
\end{equation}
where \( TL_{ref} \) is the reference trajectory length.

\noindent\textbf{\underline{D}ynamic \underline{T}ime \underline{W}arping (DTW)}: 
Given a reference path $\mathbf{R}$ and a predicted path $\mathbf{T}$, the DTW distance is defined as the minimum cumulative distance along a warping path $\omega^*$:
\[
\mathrm{DTW}(\mathbf{R}, \mathbf{T}) = \min_{\omega \in \Omega} \sum_{(i_k, j_k) \in \omega} \delta(r_{i_k}, t_{j_k}),
\]
where $\Omega$ is the set of all possible warping paths; $\delta(r_{i_k}, p_{j_k})$ is the distance function (e.g., Euclidean distance) between nodes $r_{i_k} \in \mathbf{R}$ and $t_{j_k} \in \mathbf{T}$; The optimal warping path $\omega^*$ is determined via dynamic programming to ensure minimal cumulative distance.

\noindent\textbf{\underline{N}ormalized DTW (nDTW)}: To normalize the DTW score within the range $[0, 1]$, nDTW is defined as:
\[
\mathrm{nDTW}(\mathbf{R}, \mathbf{T}) = \exp\left(-\frac{\mathrm{DTW}(\mathbf{R}, \mathbf{T})}{|\mathbf{R}| \cdot d_{\mathrm{th}}}\right),
\]
where $|\mathbf{R}|$ is the number of nodes in the reference path $\mathbf{R}$; $d_{\mathrm{th}}$ is a predefined success distance threshold. A higher nDTW score indicates a greater similarity between the predicted and reference paths. In addition, the t-nDTW is proposed by \cite{IVLN} to avoid inflating nDTW by only performing well on short tours.

\begin{table*}[t]
    \centering
    \resizebox{\textwidth}{!}{%
    \renewcommand{\arraystretch}{1}
    \setlength{\tabcolsep}{1.2mm}
    \begin{tabular}{llc|ccccccc|ccccccc}
        \toprule
        & & & \multicolumn{7}{c|}{Val-Seen} & \multicolumn{7}{c}{Val-Unseen} \\
        \cmidrule{4-10} \cmidrule{11-17}
        \# & ISM & LLM-Size
        & TL & NE$\downarrow$ & OS$\uparrow$ & nDTW$\uparrow$ & SR$\uparrow$ & SPL$\uparrow$ & t-nDTW$\uparrow$
        & TL & NE$\downarrow$ & OS$\uparrow$ & nDTW$\uparrow$ & SR$\uparrow$ & SPL$\uparrow$ & t-nDTW$\uparrow$ \\
        \midrule
        1 & \checkmark & 0.5B
        & 17.8 & \textbf{5.9} & \textbf{43} & \textbf{52} & \textbf{34} & \textbf{33} & \textbf{48}
        & 17.3 & \textbf{6.6} & \textbf{40} & \textbf{50} & \textbf{30} & \textbf{29} & \textbf{45} \\
        \midrule
        2 & \ding{53} & 0.5B
        & 17.9 & 7.1 & 37 & 46 & 28 & 23 & 42
        & 17.5 & 7.9 & 33 & 44 & 25 & 21 & 38 \\
        3 & \ding{53} & 7B
        & 18.1 & 6.4 & 40 & 49 & 31 & 27 & 45
        & 17.2 & 7.3 & 36 & 46 & 27 & 23 & 41 \\
        4 & \ding{53} & 13B
        & 17.6 & 6.2 & 41 & 50 & 32 & 29 & 46
        & 16.8 & 7.0 & 37 & 48 & 27 & 24 & 42 \\
        \bottomrule
    \end{tabular}
    }
    \caption{The ablation study for the ISM improving instruction encoding. As the number of LLM parameters increases, the instruction encoding becomes better. While ISM can significantly improve the LLM performance with a smaller parameter.}
    \label{TAB7}
    \vspace{-2mm}
\end{table*}

\begin{table*}[t]
    \centering
    \resizebox{\textwidth}{!}{%
    \renewcommand{\arraystretch}{1}
    \setlength{\tabcolsep}{1.4mm} 
    \begin{tabular}{l| ccccccc c ccccccc}
        \toprule
        & \multicolumn{7}{c}{Val-Seen} && \multicolumn{7}{c}{Val-Unseen} \\
        \cmidrule{2-8}
        \cmidrule{10-16}
        $\#$ ~Model \hspace{1em}
        & ~~~TL~~~ & NE$\downarrow$ & OS$\uparrow$ & nDTW$\uparrow$ & SR$\uparrow$ & SPL$\uparrow$ & t-nDTW$\uparrow$ 
        & ~~~ 
        & ~~~TL~~~ & NE$\downarrow$ & OS$\uparrow$ & nDTW$\uparrow$ & SR$\uparrow$ & SPL$\uparrow$ & t-nDTW$\uparrow$ \\
        \midrule
        1  ~$Maximum$ \hspace{1em}
        & ~17.5 & 6.8 & 39 & 49 & 31 & 30 & 44 
        & & 17.2 & 7.3 & 36 & 47 & 27 & 25 & 42 \\
        \midrule
        2 ~ $\Phi_\lambda= $ 0.85 \hspace{1em}
        & ~16.9 & 6.9 & 40 & 48 & 32 & 31 & 46 
        & & 17.5 & 7.2 & 38 & 48 & 27 & 26 & 43 \\
        3 ~ $\Phi_\lambda= $ 0.75 \hspace{1em}
        & ~17.1 & 6.6 & 41 & 51 & 33 & 32 & 45 
        & & 16.8 & 6.9 & 39 & 48 & 29 & 28 & 44 \\
        \rowcolor[RGB]{242,242,242}
        4 ~ $\Phi_\lambda= $ 0.65 \hspace{1em}
        & ~17.8 & \bf 5.9 & \bf 43 & \bf 52 & \bf 34 & \bf 33 & \bf 48 
        & & 17.3 & \bf 6.6 & \bf 40 & \bf 50 & \bf 30 & \bf 29 & \bf 45 \\
        5 ~ $\Phi_\lambda= $ 0.55 \hspace{1em}
        & ~17.1 & 6.2 & 41 & 50 & 32 & 31 & 46 
        & & 17.4 & 7.0 & 37 & 47 & 28 & 26 & 43 \\
        6 ~ $\Phi_\lambda= $ 0.45 \hspace{1em}
        & ~16.7 & 6.8 & 39 & 48 & 31 & 30 & 44 
        & & 17.2 & 7.4 & 37 & 46 & 27 & 25 & 42 \\
        \bottomrule
    \end{tabular}
    }
    \caption{The ablation study for \textbf{phrase selection way} and different \textbf{entropy threshold values} $\Phi_\lambda$. $Maximum$ means always selecting the phrase with the greatest similarity. We also evaluate the effect of $\Phi_\lambda$ ranging from 0.85 to 0.45 (interval = 0.1).}
    \label{TAB4}
    \vspace{-5mm}
\end{table*}

\begin{table*}[t]
    \centering
    \resizebox{\textwidth}{!}{
    \renewcommand{\arraystretch}{1}
    \setlength{\tabcolsep}{0.8mm} 
    \begin{tabular}{l| ccccccc c ccccccc}
        \toprule
        & \multicolumn{7}{c}{Val-Seen} && \multicolumn{7}{c}{Val-Unseen} \\
        \cmidrule{2-8}
        \cmidrule{10-16}
        $\#$ ~Model
        & ~~~TL~~~ & NE$\downarrow$ & OS$\uparrow$ & nDTW$\uparrow$ & SR$\uparrow$ & SPL$\uparrow$ & t-nDTW$\uparrow$
        & ~~~
        & ~~~TL~~~ & NE$\downarrow$ & OS$\uparrow$ & nDTW$\uparrow$ & SR$\uparrow$ & SPL$\uparrow$ & t-nDTW$\uparrow$ \\
        \midrule
        1 ~Type-I (w/o ISM) \hspace{1em}
        & ~16.8 & 7.6 & 36 & 47 & 28 & 23 & 43
        & & 16.4 & 8.2 & 32 & 45 & 23 & 21 & 39 \\
        2 ~Type-II  (w/ commas) \hspace{1em}
        & ~17.1 & 6.2 & 40 & 49 & 30 & 28 & 45
        & & 16.7 & 6.9 & 36 & 46 & 26 & 24 & 42 \\
        3 ~Type-III (w/ conjunctions) \hspace{1em}
        & ~16.9 & 6.4 & 39 & 48 & 29 & 28 & 44
        & & 16.7 & 7.1 & 36 & 45 & 25 & 24 & 41 \\
        \midrule
        \rowcolor[RGB]{242,242,242}
        4 ~Type-IV  (w/ periods) \hspace{1em}
        & ~17.8 & \bf 5.9 & \bf 43 & \bf 52 & \bf 34 & \bf 33 & \bf 48
        & & 17.3 & \bf 6.6 & \bf 40 & \bf 50 & \bf 30 & \bf 29 & \bf 45 \\
        \bottomrule
    \end{tabular}
    }
    \caption{The ablation study of the proposed \textbf{ISM}:  
    Type-I: without instruction segmentation.  
    Type-II: comma-based segmentation.  
    Type-III: conjunction-based segmentation.  
    Type-IV: period-based segmentation.}
    \label{TAB3}
    \vspace{-5mm}
\end{table*}

\begin{table}[t]
\centering
\renewcommand{\arraystretch}{1}
\setlength{\tabcolsep}{0.01mm}  
\begin{tabular}{cccc|cccccccc}
\toprule
\small{\#} & \small{Ver.} & \small{Term-I} & \small{Term-II} 
& TL & NE$\downarrow$ & OS$\uparrow$ & nDTW$\uparrow$ 
& SR$\uparrow$ & SPL$\uparrow$ & t-nDTW$\uparrow$ \\
\midrule
1 & \ding{53} &  &  & 13.9 & 7.0 & 37 & 46 & 29 & 26 & 43 \\
2 & \checkmark & \checkmark &  & 18.0 & 6.4 & 41 & 50 & 32 & 31 & 46 \\
3 & \checkmark &  & \checkmark & 17.6 & 6.8 & 40 & 49 & 31 & 30 & 45 \\
\rowcolor[RGB]{242,242,242}
4 & \checkmark & \checkmark & \checkmark & 17.8 & \textbf{5.9} & \textbf{43} & \textbf{52} & \textbf{34} & \textbf{33} & \textbf{48} \\
\bottomrule
\end{tabular}
\caption{Ablation study of the proposed \textbf{EaV strategy} on SH IR2R Val-seen scenes.}
\label{TAB5}
\vspace{-3mm}
\end{table}

\section{Ablation Studies}\label{AS}

The detailed ablation study results on the effects of the ISM, entropy threshold selection, instruction segmentation styles, and the terms of the EaV strategy are summarized in Table~\ref{TAB7}, Table~\ref{TAB4}, Table~\ref{TAB3}, and Table~\ref{TAB5}, based on experiments conducted on the proposed SH-IR2R scenes.  

\section{Entropy and Entropy Threshold for ISM}\label{entropyT}

This section provides further clarification on the rationale behind using entropy and setting an entropy threshold for instruction selection in ISM. 

\textit{\textbf{a). Entropy Value.}} As described in Section 4.3 of the main paper, we first compute the similarity between the $n$ phrases in the user’s instructions and the currently observed RGB image. The probability values of each phrase are then obtained through a softmax function. However, certain phrases may exhibit similar probability values due to shared object references. If we simply select the phrase with the highest probability as the final input, the agent may misinterpret the instruction, leading to errors in its decision-making process. Consequently, this could disrupt the logical execution of the instructions and result in significant deviations from the intended navigation trajectory.

For instance, consider the user’s instruction: ``Please walk straight through the table in front of you to reach the TV. Then turn left into the hallway until you see a painting on the wall. Turn right and go straight through the wooden door into the bedroom. There is a TV on your right, please stop near the table and chair opposite the TV." In this example, the objects referenced in the first phrase $\mathcal{S}^i_0$ and the last phrase $\mathcal{S}^i_n$ of the instruction exhibit significant similarities. When the agent is positioned either at the starting or ending point, the calculated probability values for these two phrases are often close. If the agent simply selects the phrase with the highest probability, it may result in biased actions, causing the agent to deviate from the logical sequence of the instructions. This deviation can lead to errors in navigation and an inability to accurately follow the intended trajectory. To mitigate this issue, we represent the similarity between the phrases in the user’s instruction by calculating the entropy value. This approach provides a comprehensive measure of uncertainty in the similarity distribution and ensures a more balanced phrase selection:
\begin{equation}
\label{model1}
\Phi^i_{t} = -\sum_{i=0}^n P(S^i_k) \log_2 P(S^i_k),
\end{equation}
where $P(S^i_k)$ denotes probability of similarity between RGB image $\mathcal{R}^i_t$ and phrases $\mathcal{S}^i_k$ in the instruction $\mathcal{I}^i$.

\textit{\textbf{b). Entropy Threshold.}} Once the agent calculates the entropy value $\Phi^i_{t}$, it compares this value against a predefined entropy threshold $\Phi_\lambda$. If $\Phi^i_{t} < \Phi_\lambda$, it indicates that the similarity between different phrases in the segmented instruction is low. In this case, the phrase with the maximum similarity, $S^i_{k^*}$, is selected as the local instruction. Conversely, if $\Phi^i_{t} \geq \Phi_\lambda$, it suggests that there are similar phrases present in the instruction. Under this condition, the agent takes the entire instruction $\mathcal{I}^i$ as the local instruction and predicts the next action based on the global context.

From Eq.~(\ref{model1}), the minimum entropy value is $0$, and the maximum value is $\log_2 n$, where $n$ is the number of phrases in the instruction. To determine an appropriate threshold $\Phi_\lambda$, we conducted experiments testing various values, ranging from $0.85$ to $0.45$ in increments of $0.1$. The results demonstrated that the model achieved optimal performance with a threshold value of approximately $0.65$. As mentioned in section 4.3 of the main paper.

\begin{figure}[t!]
\centering
\includegraphics[width=1\linewidth]{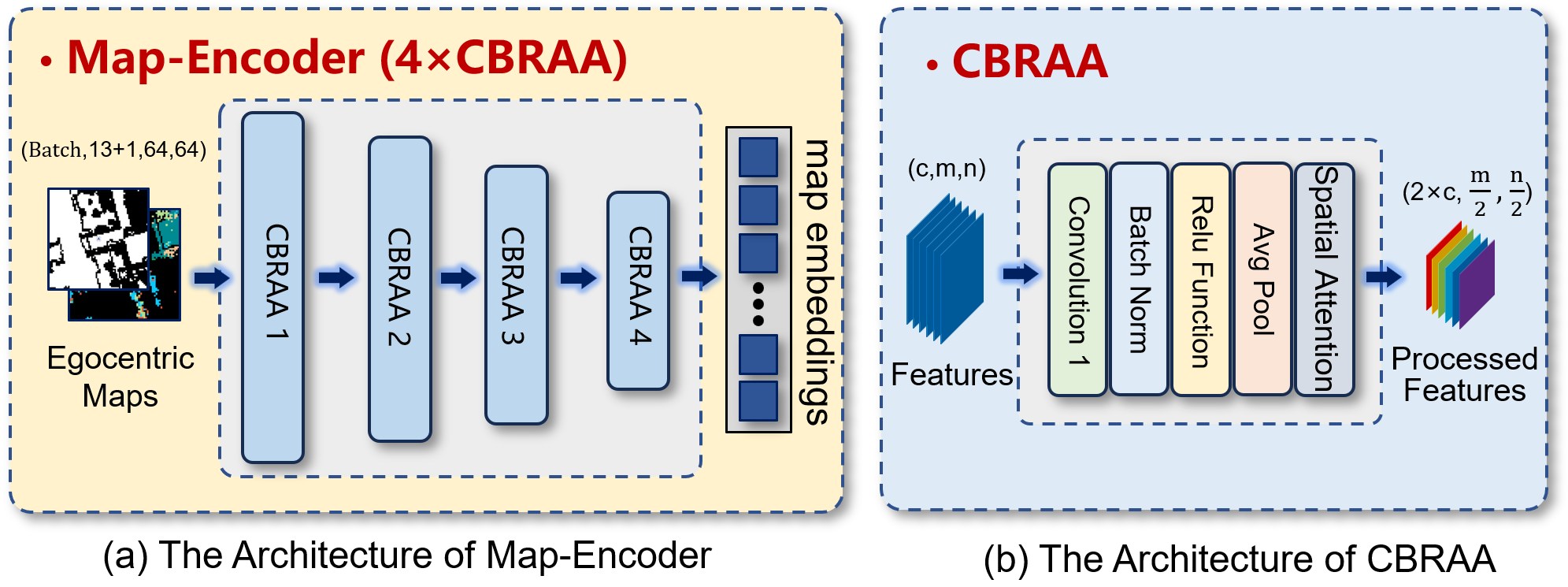}
\caption{Illustration for the architecture of the Map-Encoder.}
\label{fig_1}
\vspace{-5mm}
\end{figure}

\section{Verification Threshold}\label{VER}

In this section, we conduct an ablation study to evaluate the impact of different verification thresholds $\delta_{0}$ on the agent's performance. As described in Section 3.4, the agent determines whether it has reached the destination by calculating the similarity value. We compare the performance of learnable thresholds and fixed thresholds. The learnable similarity thresholds are based on a reward and punishment strategy. Specifically, we set the initial value of the similarity threshold $\delta_{0}$ to 0.30 in each scene, then $\delta_{0}$ decreases along with the number of steps the agent explores in the scene. For every 10 low-level actions performed by the agent, $\delta_{0}$ decreases by 0.01. If the agent verification fails once, $\delta_{0}$ decreases by 0.05. Otherwise, if the verification is successful increase the value of $\delta_{0}$ by 0.03. Notably, the minimum value of the value of the validation threshold is mandatorily set to 0.05. For each scene, the optimal threshold is obtained after several iterations. On the other hand, the fixed thresholds in this experiment start at 0.05 and gradually increase to 0.25, with a pace of 0.05. The experimental results are illustrated in Figure \ref{fig_4}. The results demonstrate that the learnable threshold $\delta_{0}$ outperforms fixed thresholds, showcasing its advantages in adapting to varying conditions. This highlights the efficacy of employing a dynamic, learnable threshold in improving the agent's ability to verify its progress and ensure accurate navigation.

\section{The Architecture of the Map-Encoder}\label{MAPEENCODER}

The Map-Encoder encodes the egocentric maps $\mathcal{M}c^i_{t}[sem]$ and $\mathcal{M}c^i_{t}[ocu]$ to feed into the CMA module. The dimension of input are $\mathcal{M}c^i_{t}[ocu]\in\mathbb{R}^{batch \times 1 \times 64 \times 64}$, $\mathcal{M}c^i_{t}[sem]\in\mathbb{R}^{batch \times 13 \times 64 \times 64}$, and the dimension of output is $map_{embedding}\in\mathbb{R}^{batch \times 128 \times 4 \times 4}$. As shown in Figure \ref{fig_1} (a), the Map-Encoder consists of four CBRAA blocks. Specificially, the extracted features pass through each CBRAA module with the number of channels multiplied by 2 and the spatial dimensions reduced to one-half of their original size. As shown in Figure \ref{fig_1} (b), each CBRAA consists of: the \textbf{\underline{C}onvolution} layer has a kernel size of 7 and does not change the number of output channels and shape size; \textbf{\underline{B}atch} normalization; \textbf{\underline{R}elu} activation function; \textbf{\underline{A}verage} pooling with has one-half pooling for shape size; Spatial \textbf{\underline{A}ttention} is used to better extract maps features. The size of the extracted features is reduced progressively to achieve efficient encoding.

\begin{figure}[t!]
\centering
\includegraphics[width=0.85\linewidth]{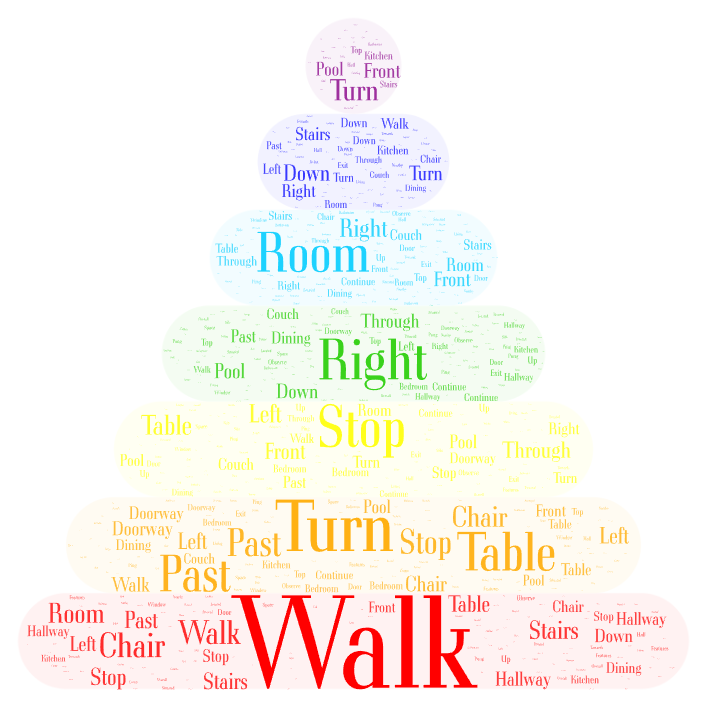}
\caption{Word clouds map of the SH IR2R-CE dataset}
\label{fig_2}
\vspace{-5mm}
\end{figure}

\section{SH IR2R-CE Dataset}\label{LSLI-IR2R-CE}

As mentioned in Section 3.3 of the main paper, we construct the \underline{L}arge-\underline{S}cene \underline{L}ong \underline{H}orizon IR2R (SH IR2R-CE) dataset to train and evaluate our SeqWalker. We enrich and stitch the existing short instructions of IR2R-CE \cite{IVLN} based on LLaVa \cite{llava} and LLaMA \cite{LLaMA}, and some manual revisions. We also provide statistical analysis of the SH IR2R-CE dataset, including comparisons of basic metrics and relevant word clouds. The word cloud map is shown in Figure \ref{fig_2}. 

Some examples of construction sequential-horizon long instructions and enrichment long instructions are listed below:

\begin{center}
\fcolorbox{black}{green!5}{\parbox{.9\linewidth}
{
\textbf{Instruction (a)}: As you walk down the hallway, turn right and look for a doorway with a mirror on it. As you walk past the stairs, turn right and continue down the hallway. Walk past the couch and stop.

\textbf{Instruction (b)}: As you walk past the couch, turn right and continue through the hallway. As you walk past the table and chairs, you will come across a large window. Stop in front of the window.

\textbf{Assistant Answer}: As you walk down the hallway, turn right and look for a doorway with a mirror on it. As you walk past the stairs, turn right and continue down the hallway. When you walk past the couch, then turn right again and continue through the hallway. As you walk past the table and chairs, you will come across a large window and wait. 
}
}
\end{center}

\begin{center}
\fcolorbox{black}{green!5}{\parbox{.9\linewidth}
{
\textbf{Instruction (a)}: As you walk past the couch, you will notice a staircase leading upwards. Please proceed up the stairs, and you will find yourself in a different area of the house. As you descend the stairs, turn left and proceed to the desk. Wait near the bed in the desk.

\textbf{Instruction (b)}: As you walk past the bed, turn right and continue down the hallway. Walk past the shelves and turn right. There is a green vase located in the middle of the scene. Stop in front of the green vase, and observe its surroundings.

\textbf{Assistant Answer}: As you walk past the couch, you will notice a staircase leading upwards. Please proceed up the stairs, and you will find yourself in a different area of the house. As you descend the stairs, turn left and proceed to the desk, where you will find the bed near the desk. Continue walking past the bed, turn right and continue down the hallway. Walk past the shelves and turn right. There is a green vase located in the middle of the scene. Stop in front of the green vase, and observe its surroundings.
}
}
\end{center}

\begin{center}
\fcolorbox{black}{blue!10}{\parbox{.9\linewidth}
{
\textbf{Instruction}: Walk through the sliding glass doors and turn left. Walk into the hallway and turn right. Walk into the bedroom and stop.

\textbf{Assistant Answer}: Walk through the sliding glass doors, which separate the living room from the outdoor patio, and turn left, heading towards the hallway. Proceed into the hallway, which leads to several rooms, and turn right, following the corridor's path. Walk into the bedroom, a space furnished with a bed, nightstands, and a dresser, and come to a halt in front of the bed.
}
}
\end{center}

\begin{center}
\fcolorbox{black}{blue!10}{\parbox{.9\linewidth}
{
\textbf{Instruction}: Walk through the doorway and turn left. Walk past the table and chairs and stop in front of the table.

\textbf{Assistant Answer}: Walk through the doorway, which separates the living area from the dining room, and turn left, heading towards the side of the room. Walk past the wooden table, surrounded by matching chairs, each with cushioned seats, and proceed further. Continue until you reach the center of the room and stop in front of the table.
}
}
\end{center}

\begin{center}
\fcolorbox{black}{blue!10}{\parbox{.9\linewidth}
{
\textbf{Instruction}: Walk past the fireplace and turn right. Walk past the fireplace and stop in front of the top of the stairs.

\textbf{Assistant Answer}: Walk past the large stone fireplace, which stands prominently in the center of the living room, and turn right, heading towards the hallway that leads to other areas of the house. Continue walking past the second fireplace, a smaller, modern design, and proceed until you reach the top of the staircase. Stop in front of the wooden banister at the top of the stairs.
}
}
\end{center}

\begin{center}
\fcolorbox{black}{blue!10}{\parbox{.9\linewidth}
{
\textbf{Instruction}: Exit the bedroom and turn right. Walk down the hallway and stop at the top of the stairs.

\textbf{Assistant Answer}: Exit the bedroom, which features a plush bed and a wardrobe, and turn right, heading towards the hallway adorned with framed pictures on the walls. Walk down the hallway, which leads to various rooms, and continue moving forward. Stop at the top of the staircase, where you will find a wooden railing and a landing that overlooks the lower floor, offering a view of the foyer and the entrance to the living room.
}
}

\end{center}

\begin{center}
\fcolorbox{black}{gray!10}{\parbox{.9\linewidth}
{
\textbf{Original instruction Type-I}: As you walk down the stairs, stop at the bottom of the stairs. Observe the surroundings and take note of the various elements in the room, such as the couch, chair, and dining table. Additionally, pay attention to the presence of a clock and a vase in the room. As you walk up the stairs, then you should stop on the third step from the top.

\textbf{Type-II}: \textcolor[rgb]{0.86, 0.078, 0.24}{As you walk down the stairs}, \textcolor[rgb]{1, 0.65, 0}{stop at the bottom of the stairs}. \textcolor[rgb]{0.5, 0.5, 0}{Observe the surroundings and take note of the various elements in the room}, \textcolor[rgb]{0.2, 0.8, 0.2}{such as the couch}, \textcolor[rgb]{0, 1, 1}{chair}, \textcolor[rgb]{0.12, 0.56, 1}{and dining table}. \textcolor[rgb]{0.54, 0.17, 0.89}{Additionally}, \textcolor[rgb]{1,0.41,0.71}{pay attention to the presence of a clock and a vase in the room}. \textcolor[rgb]{0.23, 0.66, 0.16}{As you walk up the stairs}, \textcolor[rgb]{0, 0.29, 0.22}{then you should stop on the third step from the top}.

\textbf{Type-III}: \textcolor[rgb]{0.86, 0.078, 0.24}{As you walk down the stairs, stop at the bottom of the stairs}. \textcolor[rgb]{1, 0.65, 0}{Observe the surroundings} and \textcolor[rgb]{0.5, 0.5, 0}{take note of the various elements in the room, such as the couch, chair, and dining table}. \textcolor[rgb]{0.2, 0.8, 0.2}{Additionally, pay attention to the presence of a clock } and \textcolor[rgb]{0, 1, 1}{a vase in the room}. \textcolor[rgb]{0.12, 0.56, 1}{As you walk up the stairs, }then \textcolor[rgb]{0.54, 0.17, 0.89}{you should stop on the third step from the top.}.

\textbf{Type-IV}: \textcolor[rgb]{0.86, 0.078, 0.24}{As you walk down the stairs, stop at the bottom of the stairs}. \textcolor[rgb]{1, 0.65, 0}{Observe the surroundings and take note of the various elements in the room, such as the couch, chair, and dining table}. \textcolor[rgb]{0.5, 0.5, 0}{Additionally, pay attention to the presence of a clock and a vase in the room}. \textcolor[rgb]{0.2, 0.8, 0.2}{As you walk up the stairs, then you should stop on the third step from the top}
}

}
\end{center}

\section{Different Instruction Segmentation Styles} \label{STYLES}
As shown in the ablation experiments in the main paper Section 4.3, different instruction segmentation styles have different effects. We provide an example of four segmentation styles used in the ablation experiment as follows:

\section{Navigation Visualization Examples} \label{VVLN}

\begin{figure*}[t!]
\centering
\includegraphics[width=1\linewidth]{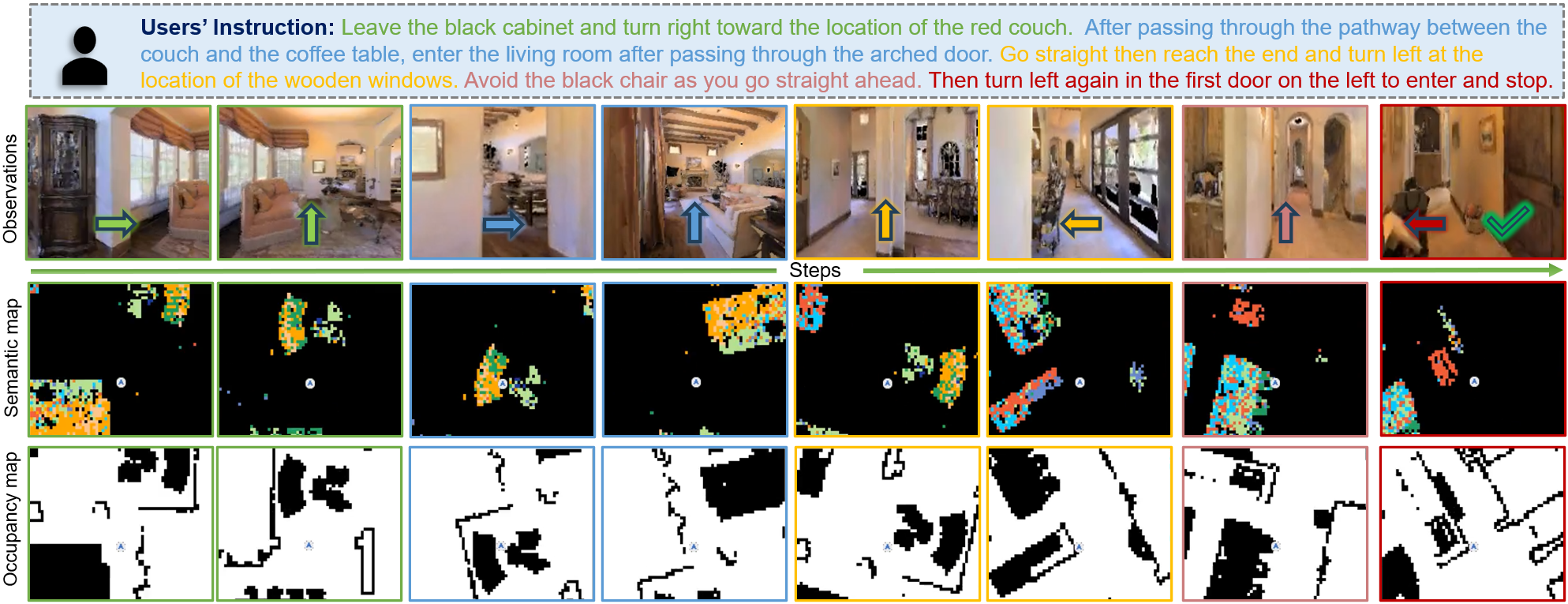}
\caption{Illustration for a visualization example of navigation in SH IR2R-CE dataset. In the top-down maps, the blue marker is the start point and the red marker is the endpoint. The agent follows the user's instructions to navigate to the goal location.}
\label{fig_5}
\end{figure*}

\begin{figure*}[t!]
\centering
\includegraphics[width=1\linewidth]{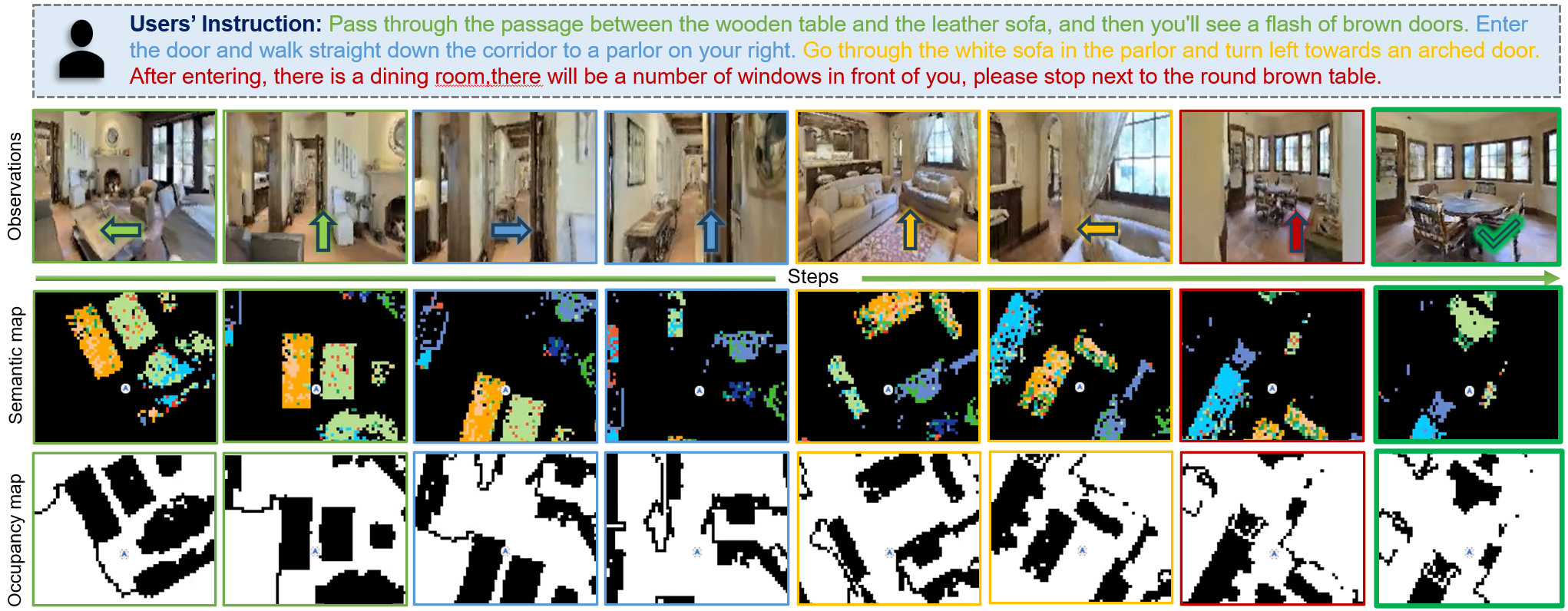}
\caption{Illustration for a visualization example of navigation in SH IR2R-CE dataset. In the top-down maps, the blue marker is the start point and the red marker is the endpoint. The agent follows the user's instructions to navigate to the goal location.}
\label{fig_6}
\end{figure*}

We provide more navigation examples to show the navigation performance of our SeqWalker, which are shown in Figure \ref{fig_5} and \ref{fig_6}. 

\section{Discussions and Future Works} \label{DaFW}

Although the proposed SeqWalker makes significant progress in indoor large-scene navigation tasks, it predominantly relies on embodied intelligent environment simulation platforms for training. While these platforms effectively simulate controlled, indoor environments, they present notable limitations when extended to open, natural scenes. Specifically, they fail to account for the complex and dynamic effects of environmental factors such as varying lighting conditions, shadows, and light scattering that are characteristic of outdoor settings. These limitations can hinder the ability of VLN models to generalize effectively to real-world outdoor navigation tasks. To solve the robust navigation problem in outdoor scenes, we provide two feasible solutions:

\begin{itemize}
    \item \textbf{Vision Enhancement Techniques}: To address the challenge of varying lighting conditions and scattering effects in natural environments, vision enhancement techniques can be employed. Approaches such as illumination normalization, contrast enhancement, and image de-noising could help the agent better perceive key visual features under changing or suboptimal lighting. These vision enhancement techniques would enhance the model's ability to interpret images from diverse conditions, improving its overall robustness.
    
    \item \textbf{Transfer Learning Strategy}: Fine-tuning the model with real-world data after training on simulations can further improve its ability to adapt to the complexities of outdoor environments. Another crucial step in enhancing generalization performance is transfer learning from real-world data. The model can acquire more robust visual features by retraining the model on diverse outdoor datasets, including data collected from outdoor sensors and cameras in various conditions (e.g., varying sunlight, scattering, rain).
\end{itemize}

\newpage

\end{document}